\documentclass[11pt]{article}

\usepackage[preprint]{acl}
\usepackage{times}
\usepackage{latexsym}
\usepackage[T1]{fontenc}
\usepackage[utf8]{inputenc}
\usepackage{microtype}
\usepackage{inconsolata}
\usepackage{graphicx}
\usepackage{amssymb,amsmath}
\usepackage{booktabs}
\usepackage{multirow}
\usepackage{colortbl}
\usepackage{hyperref}
\usepackage{tikz}
\usetikzlibrary{arrows.meta,calc,positioning,fit,backgrounds,decorations.pathreplacing}

\hypersetup{
  pdftitle={FlashMap: Flow Map Speculative Decoding},
  pdfauthor={Jeff Shen, Zhejian Zhou, Ryan Lee, Jonathan May, Shirley Ho},
}

\definecolor{oursrow}{RGB}{226,240,243}

\newcommand{\ours}{\textsc{FlashMap}}

\DeclareMathOperator{\sg}{sg}

\title{\ours{}: Flow Map Speculative Decoding}

\author{
  \textbf{Jeff Shen}\textsuperscript{1,2} \thanks{Corresponding author. \texttt{shenjeff@princeton.edu}} \quad
  \textbf{Zhejian Zhou}\textsuperscript{3} \quad
  \textbf{Ryan Lee}\textsuperscript{3} \quad
  \textbf{Jonathan May}\textsuperscript{3} \quad
  \textbf{Shirley Ho}\textsuperscript{1,2,4,5}
\\
\\
  \textsuperscript{1}Polymathic AI \quad
  \textsuperscript{2}Princeton University \\
  \textsuperscript{3}USC Information Sciences Institute \\
  \textsuperscript{4}New York University \quad
  \textsuperscript{5}Flatiron Institute
}

\begin{document}
\maketitle
\begin{abstract}
Speculative decoding accelerates large language model (LLM) inference by using a lightweight drafter to propose a block of tokens that the target model verifies in one parallel pass.
Autoregressive drafters generate this block one token at a time, whereas block diffusion drafters such as DFlash generate it in one pass, making drafting faster.
However, these diffusion drafters predict each token independently, ignoring dependencies between tokens in the block.
We introduce \ours{}, a continuous flow drafter that conditions a flow map language model on the target model's hidden states and generates the whole block in one forward pass while capturing inter-token dependencies.
On Qwen3-8B, \ours{} is faster than a matched DFlash baseline on every benchmark, with 7\% higher average speedup and up to $6.1\times$ lossless acceleration.
\\
\\
\textbf{\href{https://github.com/al-jshen/flashmap}{Code (Github)}}
\end{abstract}

\begin{figure*}[t]
  \centering
  \definecolor{tgtcol}{RGB}{49,104,166}%
\definecolor{fmcol}{RGB}{32,116,136}%
\definecolor{dfcol}{RGB}{138,134,58}%
\definecolor{badcol}{RGB}{200,40,40}%
\definecolor{nycol}{RGB}{205,55,45}%
\definecolor{sdcol}{RGB}{40,95,190}%

\providecommand{\ptcluster}[5]{%
  \foreach \n in {1,...,#3} {
    \pgfmathsetmacro{\ua}{max(rnd, 0.002)}
    \pgfmathsetmacro{\rr}{#4*sqrt(-2*ln(\ua))}
    \pgfmathsetmacro{\th}{360*rnd}
    \fill[#5] ({#1+\rr*cos(\th)}, {#2+\rr*sin(\th)}) circle (0.6pt);
  }}%

\providecommand{\xmark}[2]{%
  \fill[badcol] (#1,#2) circle (2.4pt);
  \draw[white, line width=0.7pt] (#1-0.045,#2-0.045) -- (#1+0.045,#2+0.045) (#1-0.045,#2+0.045) -- (#1+0.045,#2-0.045);}%
\begin{tikzpicture}[
  font=\footnotesize,
  >={Stealth[length=3.6pt,width=3pt]},
  line cap=round,
  mod/.style={draw, rounded corners=2pt, align=center, line width=0.5pt, inner sep=2pt},
  frozen/.style={mod, fill=black!14, draw=black!55},
  trained/.style={mod, fill=fmcol!25, draw=fmcol!85!black},
  vbox/.style={rotate=90, minimum width=1.5cm, minimum height=0.46cm},
  tok/.style={draw=black!45, fill=white, rounded corners=1.2pt, inner xsep=2pt, inner ysep=0pt,
    font=\scriptsize, minimum height=0.25cm, text height=1.5ex, text depth=0.3ex},
  arr/.style={->, line width=0.55pt, draw=black!75},
  note/.style={font=\scriptsize, align=left, text=black!80},
  lbl/.style={font=\scriptsize, inner sep=1.5pt},
]

\begin{scope}[shift={(0, 3.45)}]
  \node[font=\footnotesize\bfseries, anchor=north west, inner sep=0pt] at (0.0, 3.5) {(a)};

  \node[mod, fill=white, draw=tgtcol, minimum width=1.5cm, minimum height=1.9cm]
    (tgt) at (0.95, 1.95) {\textbf{Target}\\\textbf{LLM}\\[2pt]{\scriptsize\color{black!65}(frozen)}};
  \node[tok] (c1) at (0.42, 0.55) {we};
  \node[tok, right=1.5pt of c1] (c2) {landed};
  \node[tok, right=1.5pt of c2, draw=tgtcol, line width=0.8pt] (c3) {in};
  \draw[arr] (c2.north) -- (c2.north |- tgt.south);

  \begin{scope}[on background layer]
    \fill[fmcol!4, draw=fmcol!60, rounded corners=4pt, line width=0.6pt] (2.3, 0.25) rectangle (12.95, 3.5);
  \end{scope}
  \node[anchor=north west, font=\footnotesize\bfseries, text=fmcol!80!black] at (2.36, 3.47) {\ours{} drafter};
  \node[anchor=north east, font=\scriptsize, text=fmcol!80!black] at (12.9, 3.45)
    {all $K$ slots in one pass};

  \pgfmathsetseed{11}
  \begin{scope}[shift={(2.95, 0.85)}]
    \foreach \k in {0,...,4} {
      \foreach \v in {0,...,7} {
        \pgfmathsetmacro{\g}{12 + 62*rnd}
        \fill[black!\g] (0.15*\v, 0.3*\k) rectangle ++(0.15, 0.3);
      }
    }
    \draw[black!55, line width=0.4pt] (0,0) rectangle (1.2, 1.5);
    \node[font=\scriptsize, below=1pt] at (0.6, 0) {noise $x_0 \sim \mathcal{N}(0, I)$};
    \draw[decorate, decoration={brace, amplitude=2.5pt}, black!60] (-0.06, 0.03) -- (-0.06, 1.47)
      node[midway, left=2pt, font=\scriptsize, text=black!75] {$K$};
  \end{scope}

  \node[frozen, vbox] (emb) at (4.85, 1.6) {Embed $E$};
  \draw[arr] (4.2, 1.6) -- (emb.north);

  \node[trained, vbox] (sa) at (5.9, 1.6) {Self-Attn};
  \node[trained, vbox] (ca) at (6.85, 1.6) {Cross-Attn};
  \node[trained, vbox] (mlp) at (7.8, 1.6) {MLP};
  \draw[arr] (sa.south) -- (ca.north);
  \draw[arr] (ca.south) -- (mlp.north);

  \coordinate (dittop) at ($(ca.east)+(0,0.1)$);
  \begin{scope}[on background layer]
    \node[draw=fmcol!70, fill=white, rounded corners=3pt, line width=0.5pt, inner sep=3.5pt,
      fit=(sa)(ca)(mlp)(dittop)] (dit) {};
  \end{scope}
  \node[font=\scriptsize, anchor=south east, inner sep=1pt] at (dit.north east) {$\times L$};
  \draw[arr] (emb.south) -- (emb.south -| dit.west);

  \node[lbl] (st) at (6.85, 0.39) {adaLN$(s, t)$};
  \draw[arr] ([yshift=1.5pt]st.north) -- (dit.south -| st);

  \node[trained, minimum height=0.32cm, inner ysep=1pt] (proj) at (6.85, 2.88) {Proj};
  \draw[arr] (tgt.east |- proj) -- node[below, lbl, pos=0.5] {hidden states} (proj.west);
  \draw[arr] (proj.south) -- (ca.east);

  \node[frozen, vbox] (head) at (8.95, 1.6) {LM head};
  \draw[arr] (dit.east |- head.north) -- (head.north);

  \pgfmathsetseed{3}
  \begin{scope}[shift={(9.6, 0.85)}]
    \foreach \k/\hot in {0/6, 1/1, 2/4, 3/2, 4/5} {
      \foreach \v in {0,...,7} {
        \ifnum\v=\hot
          \fill[fmcol] (0.15*\v, 1.2-0.3*\k) rectangle ++(0.15, 0.3);
        \else
          \pgfmathsetmacro{\g}{4 + 26*rnd}
          \fill[fmcol!\g] (0.15*\v, 1.2-0.3*\k) rectangle ++(0.15, 0.3);
        \fi
      }
    }
    \draw[black!55, line width=0.4pt] (0,0) rectangle (1.2, 1.5);
    \node[font=\scriptsize, below=1pt] at (0.6, 0) {logits};
  \end{scope}
  \draw[arr] (head.south) -- (9.57, 1.6);

  \foreach \k/\w in {0/new, 1/york, 2/on, 3/friday, 4/night} {
    \node[tok, draw=fmcol!80!black, minimum width=0.82cm] (d\k) at (12.25, 2.2-0.3*\k) {\w};
  }
  \node[font=\scriptsize, above=0pt of d0] {$\hat y_{1:K}$};
  \draw[arr] (10.86, 1.6) -- node[above, lbl] {argmax} (d2.west);

  \fill[black!14, draw=black!55, line width=0.4pt, rounded corners=1pt] (13.7, 3.3) rectangle ++(0.22, 0.14);
  \node[anchor=west, font=\scriptsize, inner sep=1pt] at (13.94, 3.37) {frozen};
  \fill[fmcol!25, draw=fmcol!85!black, line width=0.4pt, rounded corners=1pt] (14.8, 3.3) rectangle ++(0.22, 0.14);
  \node[anchor=west, font=\scriptsize, inner sep=1pt] at (15.04, 3.37) {trained};

  \node[mod, fill=white, draw=tgtcol, minimum width=1.8cm, minimum height=1.9cm, font=\scriptsize]
    (ver) at (14.95, 1.6) {\footnotesize\textbf{Target}\\\footnotesize\textbf{LLM}\\[2pt]verifies\\all $K$ slots\\in one pass};
  \draw[arr] (d2.east) -- (ver.west);

  \draw[arr, tgtcol, dashed] (ver.south) -- (ver.south |- 0, 0.0) -| (0.03, 1.95) -- (tgt.west);
  \node[font=\scriptsize, text=tgtcol, fill=white, inner sep=1.5pt] at (7.6, 0.0)
    {next round: accepted tokens extend the context};
\end{scope}

\node[font=\footnotesize\bfseries, anchor=north west, inner sep=0pt] at (0.0, 3.05) {(b)};
\node[anchor=north west, font=\footnotesize, inner sep=0pt] at (0.5, 3.05)
  {Drafting two slots after \textit{we landed in}; the target continues with \textit{new york} or \textit{san diego}, each with probability $\tfrac{1}{2}$.};

\foreach \xo in {0.95, 9.0} {
  \begin{scope}[shift={(\xo, 0.45)}]
    \draw[black!25, line width=0.3pt] (0,0) rectangle (2.0,2.0);
    \foreach \i in {0.5, 1.0, 1.5} {
      \draw[black!12, line width=0.3pt] (\i,0) -- (\i,2.0);
      \draw[black!12, line width=0.3pt] (0,\i) -- (2.0,\i);
    }
    \foreach \p in {0.5, 1.0, 1.5} \foreach \q in {0.5, 1.0, 1.5} \fill[black!30] (\p,\q) circle (0.6pt);
    \node[lbl, left] at (0, 0.5) {new};
    \node[lbl, left] at (0, 1.0) {?};
    \node[lbl, left] at (0, 1.5) {san};
    \node[lbl, below] at (0.5, 0) {york};
    \node[lbl, below] at (1.0, 0) {?};
    \node[lbl, below] at (1.5, 0) {diego};
    \node[lbl, rotate=90, text=black!55] at (-0.8, 1.0) {slot 1};
    \node[lbl, text=black!55] at (1.0, -0.4) {slot 2};
  \end{scope}
}

\begin{scope}[shift={(0.95, 0.45)}]
  \foreach \tx/\ty in {0.5/0.5, 1.5/1.5, 1.5/0.5, 0.5/1.5} {
    \draw[->, dfcol!90!black, line width=0.6pt, shorten >=3.4pt, shorten <=2.2pt] (1.0,1.0) -- (\tx,\ty);
    \fill[dfcol] (\tx,\ty) circle (2.8pt);
  }
  \fill[black] (1.0, 1.0) circle (1.5pt);
  \xmark{1.66}{0.34}
  \xmark{0.34}{1.66}
\end{scope}
\node[note, anchor=west, text width=4.5cm] at (3.2, 1.45)
  {\textbf{\color{dfcol!80!black}DFlash}.
   Each slot is predicted independently, so the draft follows $q_1(y_1)\,q_2(y_2)$.
   Half of the drafts mix the modes (\textit{new diego}, \textit{san york}) and are rejected at slot~2.};

\begin{scope}[shift={(9.0, 0.45)}]
  \pgfmathsetseed{5}
  \foreach \n in {1,...,110} {
    \pgfmathsetmacro{\ua}{max(rnd, 0.002)}
    \pgfmathsetmacro{\rr}{0.15*sqrt(-2*ln(\ua))}
    \pgfmathsetmacro{\th}{360*rnd}
    \pgfmathsetmacro{\px}{\rr*cos(\th)}
    \pgfmathsetmacro{\py}{\rr*sin(\th)}
    \pgfmathtruncatemacro{\side}{(\px + \py) < 0}
    \ifnum\side=1
      \fill[nycol, opacity=0.25] (1.0+\px, 1.0+\py) circle (0.6pt);
    \else
      \fill[sdcol, opacity=0.25] (1.0+\px, 1.0+\py) circle (0.6pt);
    \fi
  }
  \draw[->, nycol, line width=0.7pt, shorten >=3.6pt] (0.8, 0.8) -- (0.5, 0.5);
  \draw[->, sdcol, line width=0.7pt, shorten >=3.6pt] (1.2, 1.2) -- (1.5, 1.5);
  \fill[nycol] (0.5, 0.5) circle (2.8pt);
  \fill[sdcol] (1.5, 1.5) circle (2.8pt);
\end{scope}
\node[note, anchor=west, text width=4.5cm] at (11.18, 1.45)
  {\textbf{\color{fmcol}\ours{}}.
   Each noise sample $x_0$ is sent to a single block $X_{0,1}(x_0)$, so the noise selects the mode for both slots jointly.
   Every draft is \textit{new york} or \textit{san diego}.};

\draw[black!20, line width=0.4pt] (8.0, 0.1) -- (8.0, 2.55);

\end{tikzpicture}%
\par
  \vspace{-2pt}
  {\raggedright\footnotesize\textbf{(c)}\par}
  \vspace{-5pt}
  {\small
  \setlength{\tabcolsep}{3.5pt}
  \begin{tabular}{@{}llccccccccc@{}}
    & & \multicolumn{3}{c}{Math} & \multicolumn{3}{c}{Code} & \multicolumn{3}{c}{Chat} \\
    \cmidrule(lr){3-5} \cmidrule(lr){6-8} \cmidrule(l){9-11}
    Temp. & Method & GSM8K & MATH & AIME25 & MBPP & HumanEval & LCB & MT-Bench & Alpaca & Arena-Hard \\
    \midrule
    \multirow{4}{*}{$T=0$}
      & EAGLE-3 (16) & 3.20 & 3.09 & 2.82 & 2.57 & 2.87 & 2.90 & 1.80 & 1.78 & 1.92 \\
      & DFlash & 5.77 & 5.59 & 4.67 & 3.74 & 4.50 & 4.88 & 2.48 & 2.34 & 2.67 \\
      & \cellcolor{oursrow}\ours{} (1 step) & \cellcolor{oursrow}\textbf{6.14} & \cellcolor{oursrow}\textbf{5.89} & \cellcolor{oursrow}\textbf{4.89} & \cellcolor{oursrow}\textbf{4.02} & \cellcolor{oursrow}\textbf{4.73} & \cellcolor{oursrow}\textbf{5.26} & \cellcolor{oursrow}\textbf{2.73} & \cellcolor{oursrow}\textbf{2.56} & \cellcolor{oursrow}\textbf{3.04} \\
      & \cellcolor{oursrow}\ours{} (2 steps) & \cellcolor{oursrow}5.89 & \cellcolor{oursrow}5.57 & \cellcolor{oursrow}4.61 & \cellcolor{oursrow}3.79 & \cellcolor{oursrow}4.53 & \cellcolor{oursrow}5.06 & \cellcolor{oursrow}2.47 & \cellcolor{oursrow}2.33 & \cellcolor{oursrow}2.76 \\
    \addlinespace
    \multirow{4}{*}{$T=1$}
      & EAGLE-3 (16) & 2.94 & 2.69 & 2.24 & 2.37 & 2.53 & 2.51 & 1.66 & 1.57 & 1.55 \\
      & DFlash & 4.89 & 4.24 & 3.23 & 3.42 & 3.82 & 3.54 & 2.24 & 2.06 & 2.05 \\
      & \cellcolor{oursrow}\ours{} (1 step) & \cellcolor{oursrow}\textbf{5.12} & \cellcolor{oursrow}\textbf{4.39} & \cellcolor{oursrow}\textbf{3.56} & \cellcolor{oursrow}\textbf{3.47} & \cellcolor{oursrow}\textbf{3.95} & \cellcolor{oursrow}\textbf{3.85} & \cellcolor{oursrow}\textbf{2.44} & \cellcolor{oursrow}\textbf{2.28} & \cellcolor{oursrow}\textbf{2.30} \\
      & \cellcolor{oursrow}\ours{} (2 steps) & \cellcolor{oursrow}5.07 & \cellcolor{oursrow}4.15 & \cellcolor{oursrow}3.28 & \cellcolor{oursrow}3.43 & \cellcolor{oursrow}3.86 & \cellcolor{oursrow}3.57 & \cellcolor{oursrow}2.29 & \cellcolor{oursrow}2.14 & \cellcolor{oursrow}2.14 \\
    \bottomrule
  \end{tabular}}
  \caption{
    (a)~One round of speculative decoding: the drafter reads hidden features from the target and proposes the next $K$ tokens in parallel with a flow map, and the target verifies them in one pass.
    (b)~Sampling two correlated slots in the few-step regime (adapted from Figure~3 of FMLM): DFlash predicts each slot independently, so half of its drafts pair tokens from different modes, while \ours{} can represent a joint over all slots.
    \ (c)~Speedup ($\times$) on Qwen3-8B, with \ours{} drafting in 1 or 2 flow-map steps. \textbf{Bold} marks the best speedup.
  }
  \label{fig:overview}
\end{figure*}

\section{Introduction}
Speculative decoding drafts a block of future tokens with a lightweight model and verifies it in one parallel pass of the larger target LLM. As a result, speedup depends on how many drafted tokens are accepted and how cheaply, in terms of compute, they are drafted.
EAGLE-3~\citep{li_eagle-3_2025} achieves high acceptance by conditioning on the hidden states of the target model but drafts autoregressively so cost grows with block length.

DFlash~\citep{chen_dflash_2026} instead drafts an entire block in one forward pass using a lightweight block diffusion model, yielding over 6$\times$ lossless acceleration.
However, a single masked-denoising pass predicts each position in the block independently.
Prior work finds that draft tokens from this model can be individually plausible but jointly inconsistent, leading to smaller accepted prefixes, and acceptance decays at later positions~\citep{cheng_dspark_2026}. 

By changing the drafter's generative model, we no longer draft each token independently (known as the \emph{factorized approximation}).
Flow Map Language Models~\citep{lee_flow_2026} distill a continuous flow over one-hot token embeddings into a flow map that transports Gaussian noise to a full sequence in one network evaluation.
Because the map acts on the whole block jointly, it can represent dependencies between tokens that a drafter predicting each position independently cannot, while keeping DFlash's one-pass cost (Figure~\ref{fig:overview}).

\paragraph{Contributions.}
We evaluate \ours{} on Qwen3-8B across nine math, code, and chat benchmarks (Qwen3-4B in Appendix~\ref{sec:appendix-4b}), against EAGLE-3 and DFlash retrained under identical data and settings.
Our contributions are:

\vspace{4pt}
\noindent\colorbox{oursrow}{\parbox{\dimexpr\linewidth-2\fboxsep\relax}{%
(1)~\textbf{Flow map drafting.} We introduce \ours{}, the first speculative decoding drafter built on a continuous flow. \ours{} combines DFlash's target-conditioning, single-pass block drafting with flow map language models, allowing it to draft a whole block that carries inter-token dependencies in a way that DFlash cannot.

\par\smallskip
(2)~\textbf{Analysis of flow map drafts.} We study what drives \ours{}'s acceptance length, using proposal-length ablations, conditional acceptance curves, and a shuffle test that separates the contribution of per-slot predictions from that of inter-token dependencies.%
}}

\section{Background}

\paragraph{Speculative decoding.}
Let $p$ be an autoregressive target model over a vocabulary $\mathcal{V}$ and $c$ the current context.
In each round, a drafter proposes a block $\hat y_{1:K}$ from a proposal $q(\cdot \mid c)$, and the target scores all $K$ positions in one forward pass.
The target accepts tokens left to right with the rejection rule of \citet{leviathan_fast_2023}, which keeps the output distributed exactly as $p$.
At the first rejection, the target resamples that token from the residual distribution and discards the rest of the block.
Under greedy decoding ($T=0$), the rule reduces to keeping the longest prefix that matches the target's argmax.
The acceptance length $\gamma \in [1, K+1]$ is the expected number of tokens committed per round, including the one contributed by the target.
The per-token latency is $(T_{\text{draft}} + T_{\text{verify}})/\gamma$~\citep{chen_dflash_2026}, where $T_{\text{verify}}$ is roughly constant because decoding is memory-bound~\citep{chen_accelerating_2023}.

\paragraph{DFlash.}
DFlash~\citep{chen_dflash_2026} drafts with a block discrete diffusion model.
It attends to hidden states from several target layers and denoises a fully masked block in a single forward pass, so $T_{\text{draft}}$ is independent of $K$.
Its token embedding and language modeling (LM) head are copied from the target and kept frozen, so only a few transformer layers are trained.
Training samples random anchor positions in target-generated responses and weights the per-slot cross-entropy by exponentially decaying weights over block positions.

\paragraph{Factorization error.}
Discrete diffusion samples tokens at every denoising step.
Sampling several tokens in one step would require their joint distribution over $\mathcal{V}^K$, whose size is exponential in $K$ and thus intractable, so few-step inference with discrete diffusion typically relies on the factorized approximation: each position is instead predicted independently given the noisy state~\citep{lee_flow_2026}.
A one-pass drafter, whose input is the fully masked block, therefore proposes
\begin{equation}
  q(y_{1:K} \mid c) = \prod_{k=1}^{K} q_k(y_k \mid c).
  \label{eq:factorized}
\end{equation}
Per-slot cross-entropy is minimized by the target's marginals $q_k = p(y_k \mid c)$.
The remaining gap is the total correlation $\mathrm{KL}\big(p(y_{1:K} \mid c) \,\|\, \prod_k p(y_k \mid c)\big)$, which no amount of model capacity removes.
For example, if the continuation is ``new york'' or ``san diego'' with equal probability, the drafter proposes ``new diego'' or ``san york'' half of the time~\citep{lee_flow_2026}.
This is a problem that plagues many non-autoregressive methods \citep[e.g.,][]{gu2018nonautoregressiveneuralmachinetranslation}.
Taking a large number of steps at inference time recovers these dependencies, but is antithetical to the goal of a fast, few-pass drafter.

\paragraph{Flows over one-hot tokens.}
Flow Map Language Models (FMLM;~\citealp{lee_flow_2026}) instead model language with a continuous flow.
The flow's only randomness is the initial noise, which a deterministic trajectory transports to the whole block jointly.
A sequence $y \in \mathcal{V}^L$ is represented by its one-hot encoding $x_1 \in \mathbb{R}^{L \times |\mathcal{V}|}$ and decoded with an argmax.
Following rectified flow \citep{lipman_flow_2023,liu2022rf,albergo_stochastic_2025}, noise and data are connected by
\begin{equation}
  I_t = (1-t)\,x_0 + t\,x_1, \qquad x_0 \sim \mathcal{N}(0, I),
  \label{eq:interpolant}
\end{equation}
and the velocity $v_t(x) = \mathbb{E}[x_1 - x_0 \mid I_t = x]$ transports $x_0$ at $t=0$ to data at $t=1$.
FMLM predicts the clean data rather than the velocity, since the clean data are low-rank while the velocity is noisy and full-rank~\citep{li2026basics}.
A denoiser $D^\theta_t$ outputs the clean-data prediction, and the velocity is recovered as
\begin{equation}
  v^\theta_t(x) = \frac{D^\theta_t(x) - x}{1 - t}.
  \label{eq:velocity}
\end{equation}
Because one-hot vectors are the vertices of the simplex, the true denoiser $D_t(x)^l = p(y_l = \cdot \mid I_t = x)$ lies on the simplex at every position.
FMLM therefore parameterizes $D^\theta_t$ with a softmax and trains it with cross-entropy,
\begin{equation}
  \mathcal{L}_{\mathrm{CE}}(\theta) = \mathbb{E}_{t, x_0, x_1}\Big[-\sum_{l=1}^{L} x_1^l \cdot \log D^\theta_t(I_t)^l\Big],
  \label{eq:flm_ce}
\end{equation}
whose unique minimizer is the true denoiser.
Sampling integrates the flow with a numerical ODE solver, typically over tens of steps.

\paragraph{Flow maps.}
The flow map $X_{s,t}$ sends the ODE's state at time $s$ to its state at time $t$~\citep{boffi2025flowmap,boffi2025consistency}.
It satisfies $X_{s,s} = \mathrm{id}$ and the semigroup property $X_{u,t} \circ X_{s,u} = X_{s,t}$.
FMLM parameterizes it through a two-time denoiser $\delta_{s,t}$ on the simplex,
\begin{equation}
  X_{s,t}(x) = \frac{1-t}{1-s}\,x + \frac{t-s}{1-s}\,\delta_{s,t}(x),
  \label{eq:two_time}
\end{equation}
which reduces to the denoiser on the diagonal, $\delta_{s,s} = D_s$~\citep[Prop.~3.3]{lee_flow_2026}.
A single evaluation $X_{0,1}(x_0) = \delta_{0,1}(x_0)$ thus maps noise to data.
The map $\delta$ is obtained by distilling a pretrained flow (\S\ref{sec:distill}).

\section{Method}

Our drafter is a flow over one-hot draft blocks, conditioned on the target in the same way as DFlash.
We first train it as a flow (\S\ref{sec:flow}) and then distill it into a two-time denoiser that drafts a whole block in a single network evaluation (\S\ref{sec:distill}).
Let $a$ be the position of the anchor $y_a$, the last token committed by the target.
The drafter proposes the next $K$ tokens, and we index the block by slot $k = 1, \dots, K$, so that slot $k$ holds $y_{a+k}$.
The drafter operates on a noisy block, which we call the \emph{state} $x \in \mathbb{R}^{K \times |\mathcal{V}|}$: the point on the flow's trajectory between Gaussian noise ($t = 0$) and the one-hot encoding of the $K$ draft tokens ($t = 1$).
During training the state is the interpolant $I_t$, and during drafting it is the current iterate of the flow.
We write $x^k \in \mathbb{R}^{|\mathcal{V}|}$ for the state at slot $k$, and $c$ for the target features available at drafting time.

\subsection{Continuous Flow Drafter}
\label{sec:flow}

The drafter's denoiser $D^\theta_t(x; c)$ is trained as in FMLM (Eqs.~\ref{eq:interpolant}--\ref{eq:flm_ce}), with changes to the architecture, the input featurization, and the loss weighting.

\paragraph{Architecture.}
The backbone is a conditional diffusion transformer (DiT;~\citealp{peebles_scalable_2023}).
Each block applies self-attention among the $K$ slots, cross-attention to context features from the target, and an MLP, each modulated by adaLN-Zero on the time $t$.
The context features are the target's hidden states from a fixed set of layers, concatenated and linearly projected at each position before the anchor, together with an embedding of the anchor token $y_a$.
Both vocabulary projections are copied from the target and kept frozen: the state enters the network through the target's embedding matrix $E \in \mathbb{R}^{|\mathcal{V}| \times d}$, and the output passes through the target's LM head, with a softmax over the vocabulary at each slot giving $D^\theta_t(x; c)$.

\paragraph{Training.}
We train on target-generated responses with cached target features and sample many anchor positions per sequence, as in DFlash.
Each block receives its own time $t$ and noise $x_0$ and is noised to $I_t = (1-t)\,x_0 + t\cdot\mathrm{onehot}(y_{a+1:a+K})$.
Following DFlash, we weight the cross-entropy of Eq.~\eqref{eq:flm_ce} at slot $k$ by $w_k = e^{-(k-1)/\kappa}$.
The weights emphasize early slots, since an early error invalidates every later token.

\paragraph{Time reparameterization.}
With a large vocabulary (e.g., $|\mathcal{V}| = 151{,}936$ for Qwen3), the state reveals the clean tokens only within a narrow window of $t$ near $1$ \citep{lee_flow_2026}.
Sampling $t$ uniformly therefore spends most of training at times where the state reveals almost nothing or has already revealed everything.
We adopt FMLM's decoding-error reparameterization, which measures time by how many tokens have been resolved.
Let $P_e(t)$ be the probability that decoding the state at a slot by its argmax gives the wrong token; it falls from $1 - 1/|\mathcal{V}|$ under pure noise to $0$ at $t = 1$.
The reparameterized time $\tau(t) = 1 - P_e(t)/P_e(0)$ is the fraction of this initial error removed by time $t$, so equal steps in $\tau$ make equal progress in resolving tokens.
For our interpolant, $P_e(t)$ has the closed form
\begin{equation}
  P_e(t) = 1 - \mathbb{E}_{z \sim \mathcal{N}(0,1)}\Big[\Phi\big(z + \tfrac{t}{1-t}\big)^{|\mathcal{V}|-1}\Big],
  \label{eq:pe}
\end{equation}
where $\Phi$ is the standard normal CDF.
We sample $\tau \sim \mathcal{U}[0,1]$ during training and map it back to $t$ by inverting $\tau(t)$; at inference, we use the grid $t_n = \tau^{-1}(n/N)$.
Drafting directly from the flow takes $N$ Euler steps on this grid, and thus costs $N$ NFEs per drafting round, typically $\mathcal{O}(100)$.

\subsection{Distillation into a Flow Map}
\label{sec:distill}

Drafting with $\mathcal{O}(100)$ network evaluations per round is expensive, and taking fewer Euler steps reduces this cost at the price of a less accurate draft.
In the extreme of a single step from $t = 0$, the draft loses all joint structure: it is the denoiser's prediction from pure noise, which is trained on noise drawn independently of the data.
Its best value is therefore the per-slot marginals whatever the noise, i.e., the factorized drafter of Eq.~\eqref{eq:factorized}.
The noise shapes the sample only through integration, as each step moves the state in a noise-dependent direction that later evaluations resolve into one coherent block.

A flow map~\citep{boffi2025flowmap,boffi2025consistency} is the solution operator of the flow's ODE: $X_{s,t}$ sends the state at any time $s$ to the state the ODE reaches at any later time $t$.
Learning it replaces numerical integration over $[s, t]$ with a single network evaluation, so the exact map preserves the flow's joint distribution over the block at any number of jumps, including one.
In practice, a learned map approximates long jumps less accurately than short ones, so at 1-NFE it captures this joint structure only partially (\S\ref{sec:shuffle}).
Following the flow map distillation framework of \citet{boffi2025flowmap}, we distill the trained flow (the teacher) into a flow map (the student), a two-time denoiser $\delta^\phi_{s,t}(x; c)$ that estimates the jump from $s$ to $t$.
With a single jump, the draft is $X_{0,1}(x_0) = \delta^\phi_{0,1}(x_0; c)$; more jumps can be composed at the cost of one evaluation each~\citep{boffi2025flowmap}.

\paragraph{Semigroup distillation.}
We train the student on the semigroup property of flow maps~\citep{boffi2025flowmap,boffi2025consistency}: a jump from $s$ to $t$ must land in the same place as a jump from $s$ to an intermediate time $u$ followed by a jump from $u$ to $t$.
For zero-length jumps ($s = t$), the target is the frozen flow's denoiser $D^\theta_s(I_s)$, which enforces the tangent condition $\delta_{s,s} = D_s$~\citep{boffi2025consistency} and anchors the student to the teacher.
For longer jumps ($s < t$), the target is the student's own estimate from splitting the jump at the midpoint $u$, which in terms of the two-time denoiser is a weighted average~\citep{lee_flow_2026}:
\begin{equation}
  \bar\delta_{s,t} = \sg\Big(\lambda\,\delta^\phi_{s,u}(I_s) + (1-\lambda)\,\delta^\phi_{u,t}\big(X^\phi_{s,u}(I_s)\big)\Big),
  \label{eq:semigroup}
\end{equation}
where $\lambda = \frac{(1-t)(u-s)}{(1-u)(t-s)}$, $X^\phi_{s,u}$ is the student's first jump (Eq.~\ref{eq:two_time}), and $\sg$ denotes the stop-gradient, since the target is computed from the student itself.
Training toward this target bootstraps from short jumps, which are anchored to the teacher, to progressively longer ones~\citep{boffi2025consistency}.
Following FMLM, half of the blocks train zero-length jumps and the rest train jumps of random start and length, including a small fraction of full jumps $(0,1)$.
The resulting map is valid for any pair of times, so it can also draft with several shorter jumps.
Appendix~\ref{sec:appendix-fast} describes an optimization for multi-step jumps.

\begin{table*}[!t]
  \centering
  \small
  \setlength{\tabcolsep}{3.5pt}
  \begin{tabular}{@{}llccccccccc@{}}
    \toprule
    \multirow{2}{*}{Temp.} & \multirow{2}{*}{Method} & \multicolumn{3}{c}{Math} & \multicolumn{3}{c}{Code} & \multicolumn{3}{c}{Chat} \\
    \cmidrule(lr){3-5} \cmidrule(lr){6-8} \cmidrule(l){9-11}
    & & GSM8K & MATH & AIME25 & MBPP & HumanEval & LCB & MT-Bench & Alpaca & Arena-Hard \\
    \midrule
    \multirow{4}{*}{$T=0$}
      & EAGLE-3 (16) & 6.75 & 6.43 & 5.96 & \textbf{5.38} & 6.04 & 6.16 & \textbf{3.75} & \textbf{3.71} & \textbf{3.97} \\
      & DFlash & 7.43 & 7.09 & 5.92 & 4.77 & 5.74 & 6.35 & 3.14 & 2.96 & 3.38 \\
      & \cellcolor{oursrow}\ours{} (1 step) & \cellcolor{oursrow}7.69 & \cellcolor{oursrow}7.24 & \cellcolor{oursrow}6.00 & \cellcolor{oursrow}4.95 & \cellcolor{oursrow}5.82 & \cellcolor{oursrow}6.65 & \cellcolor{oursrow}3.34 & \cellcolor{oursrow}3.14 & \cellcolor{oursrow}3.71 \\
      & \cellcolor{oursrow}\ours{} (2 steps) & \cellcolor{oursrow}\textbf{8.31} & \cellcolor{oursrow}\textbf{7.69} & \cellcolor{oursrow}\textbf{6.36} & \cellcolor{oursrow}5.24 & \cellcolor{oursrow}\textbf{6.26} & \cellcolor{oursrow}\textbf{7.19} & \cellcolor{oursrow}3.40 & \cellcolor{oursrow}3.20 & \cellcolor{oursrow}3.80 \\
    \addlinespace
    \multirow{4}{*}{$T=1$}
      & EAGLE-3 (16) & 6.15 & 5.55 & \textbf{4.61} & \textbf{4.94} & 5.29 & \textbf{5.37} & \textbf{3.43} & \textbf{3.25} & \textbf{3.20} \\
      & DFlash & 6.30 & 5.40 & 4.10 & 4.37 & 4.88 & 4.62 & 2.85 & 2.62 & 2.61 \\
      & \cellcolor{oursrow}\ours{} (1 step) & \cellcolor{oursrow}6.40 & \cellcolor{oursrow}5.40 & \cellcolor{oursrow}4.37 & \cellcolor{oursrow}4.29 & \cellcolor{oursrow}4.87 & \cellcolor{oursrow}4.90 & \cellcolor{oursrow}3.00 & \cellcolor{oursrow}2.81 & \cellcolor{oursrow}2.83 \\
      & \cellcolor{oursrow}\ours{} (2 steps) & \cellcolor{oursrow}\textbf{7.11} & \cellcolor{oursrow}\textbf{5.74} & \cellcolor{oursrow}4.52 & \cellcolor{oursrow}4.75 & \cellcolor{oursrow}\textbf{5.35} & \cellcolor{oursrow}5.06 & \cellcolor{oursrow}3.16 & \cellcolor{oursrow}2.96 & \cellcolor{oursrow}2.95 \\
    \bottomrule
  \end{tabular}
  \caption{Acceptance length $\gamma$ on Qwen3-8B.
  For each temperature and benchmark, bold marks the best result. Speedups are in Figure~\ref{fig:overview}(c).}
  \label{tab:acceptance}
\end{table*}

\section{Experimental Setup}
\label{sec:setup}

\paragraph{Models.}
We use Qwen3-4B and Qwen3-8B \citep{yang2025qwen3technicalreport} as target models and implement all drafters in DeepSpec\footnote{\url{https://github.com/deepseek-ai/DeepSpec}}.
We extract the target model's hidden states at layers 1, 9, 17, 25, and 33, to be used as inputs to all drafters.

\ours{} uses 6 DiT blocks \citep{peebles_scalable_2023}, embedding dimension 2048, 16 attention heads, QK-norm, and GELU activations in the MLPs.
We note that DFlash uses the same embedding dimension as the target model.
Thus, for Qwen3-8B, DFlash's embedding dimension is 4096, and the drafter has a total of ${\sim}$1B parameters, while \ours{} has ${\sim}$0.5B parameters.
For EAGLE-3 we use training-time test of TTT=7, top-k of 10, and a tree size of 16 to ensure that all methods have the same verification budget. 

\paragraph{Training.}
We train on 500K samples from Open-PerfectBlend\footnote{\url{https://huggingface.co/datasets/mlabonne/open-perfectblend}}, an open-source version of PerfectBlend \citep{xu2024perfectblendredefiningrlhf}; we use only a subset of the full dataset because of storage and compute limits.
We train with block size $K = 16$, and set the slot loss decay $w_k$ to $\kappa = 7$.
We use the AdamW optimizer \citep{loshchilov2019decoupledweightdecayregularization} with $(\beta_1, \beta_2) = (0.9, 0.95)$ and weight decay $0.1$, which we find important for stable training.
We use a cosine learning rate schedule with 4\% warmup and a peak learning rate of $6 \times 10^{-4}$.
We train in bf16 with gradient-norm clipping at $1.0$ and a global batch size of 256 for 20 epochs (39K steps).
The EAGLE-3 and DFlash baselines are retrained with the same data and under the same settings for controlled comparison.

\paragraph{Distillation.}
We initialize the flow map from the weights of the flow teacher.
A fraction $1/16$ of the jumps are full jumps $(s, t) = (0, 1)$, and we use a warmup-stable-decay learning rate schedule with 4\% warmup, peak learning rate of $5\times10^{-5}$, and 20\% decay.
We use no weight decay and train for 12 epochs (23k steps).
All other settings match those of the teacher.

\paragraph{Evaluation.}
We evaluate on nine benchmarks in three domains: math (GSM8K \citep{cobbe2021trainingverifierssolvemath}, MATH \citep{hendrycks2021measuringmathematicalproblemsolving}, and AIME25 \citep{aime25}), code (MBPP \citep{austin2021programsynthesislargelanguage}, HumanEval \citep{chen2021evaluating}, and LiveCodeBench (LCB; \citet{jain2024livecodebenchholisticcontaminationfree})), and chat (MT-Bench \citep{Bai_2024}, Alpaca \citep{alpaca_eval}, and Arena-Hard \citep{li2024crowdsourceddatahighqualitybenchmarks}).
We report the acceptance length $\gamma$ and the end-to-end speedup over autoregressive decoding with the target, at temperatures $T = 0$ and $T = 1$.
\ours{} drafts with one or two steps, i.e., jumps of the flow map.
All evaluations use SGLang at concurrency 1 and speedups are measured on an NVIDIA H200 NVL GPU, unless otherwise specified. 

\begin{figure*}[t]
  \centering
  \includegraphics[width=\textwidth]{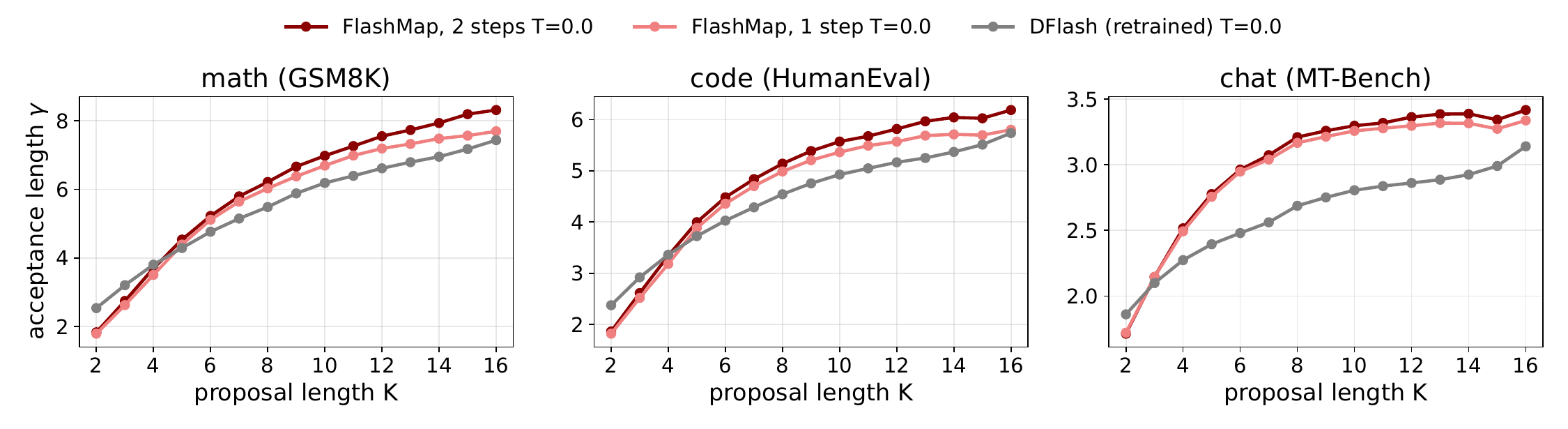}
  \caption{Acceptance length $\gamma$ against the proposal length $K$ used at inference on Qwen3-8B at $T=0$, for \ours{} with one and two steps and for DFlash, retrained with the same data and settings, on GSM8K, HumanEval, and MT-Bench (80 prompts each).
  All drafters were trained at $K=16$ and run at smaller $K$ without retraining.
  Each point is a single run, so no uncertainty is shown.}
  \label{fig:proplen8b}
\end{figure*}

\section{Results}

We show the benefit of speculative decoding methods over plain autoregressive decoding with the target model in terms of speedup (Figure~\ref{fig:overview}(c)) and acceptance length $\gamma$ (Table~\ref{tab:acceptance}) for various draft models, as applied to Qwen3-8B.
We show analogous results for Qwen3-4B in Appendix~\ref{sec:appendix-4b}.
Across all models, speedups are largest for math and code tasks, where text is highly structured, and smaller for chat tasks, which are more open-ended;
in chat, a given piece of text can have many possible relevant continuations.

\paragraph{One step \ours{} is the fastest drafter.}
At both $T=0$ and $T=1$, \ours{} at 1 NFE leads to the highest speedup over the base model across every task and domain, with speedups of up to $6.14\times$ for GSM8K at $T=0$.
As compared to DFlash, \ours{} at 1 NFE is $7.4\%$ faster on average across all tasks.

\paragraph{One step \ours{} also drafts more accurately than DFlash.}
Although \ours{} at 1 NFE does not have the highest acceptance length in all, or even a majority of the tasks, it does have higher acceptance length than DFlash in 15/18 settings, despite having half the parameters of DFlash.
This indicates that the wins in speedup are not coming from a smaller model alone (i.e., reduced latency)---\ours{} at 1 NFE is also a highly accurate model.
Our gains over DFlash are largest in the chat tasks---the most difficult ones.
In particular, at $T=0$, while acceptance length/speedup on math tasks is only 2.3\%/5.5\% higher than for DFlash, on chat tasks this increases to 7.4\%/11.1\%.

\paragraph{AR drafters are accurate but slow.}
On the acceptance length front, \ours{} at 2 NFE and EAGLE-3 are both stronger than \ours{} at 1 NFE, especially at $T=1$.
In particular, at $T=1$, EAGLE-3 is the drafter with the highest acceptance length in 6/9 tasks.
However, EAGLE-3, being an autoregressive drafter, requires multiple sequential passes to produce a single draft, resulting in very high latency.
Given that total speedup depends both on acceptance length and latency, the small edge that EAGLE-3 has in terms of acceptance length does not offset its higher latency, leading to EAGLE-3 being not competitive with parallel drafters.
Indeed, at $T=1$, averaged across tasks, EAGLE-3 has 9\% higher acceptance length but is a whopping 35\% slower than \ours{} at 1 NFE.

\paragraph{Two step \ours{} has the best acceptance of any parallel drafter.}
\ours{} at 2 NFE beats DFlash in acceptance length by an average of $10\%$, and wins 18/18 settings (at $T=0$ and $T=1$).
It also remains slightly faster than DFlash, winning 13/18 settings, despite two draft passes (Appendix~\ref{sec:appendix-fast}).

As an alternative to more function evaluations, another way to spend more drafting compute is to draft several blocks from independent noise draws and verify them together as a tree, which we explore in Table~\ref{tab:multisample};
we find that this can lead to speedups of ${\sim}11\%$ over single-sample drafting with standard chain verification.

\begin{figure*}[t]
  \centering
  \includegraphics[width=\textwidth]{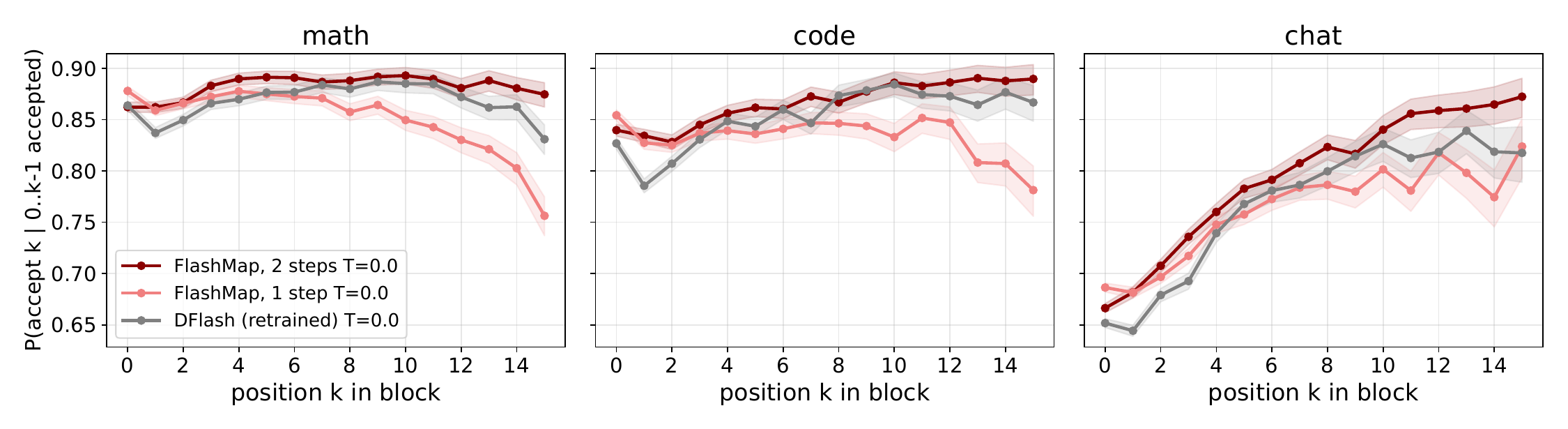}
  \caption{Conditional acceptance rate on Qwen3-8B at $T=0$ for the drafters of Figure~\ref{fig:proplen8b}: the probability that the draft token at position $k$ of a $K=16$ block is accepted given that all earlier tokens in the block were accepted, pooled over the math, code, and chat tasks.
  Positions are 0-indexed, so position $k$ is slot $k+1$; shaded bands are 95\% Wilson intervals.}
  \label{fig:decay8b}
\end{figure*}

\section{Discussion}
\label{sec:discussion}

The results above show that \ours{} at one step is the fastest drafter we evaluate.
We now ask where its gains come from, whether its drafts use dependencies between tokens, and what limits one-step drafting.

\subsection{Where is acceptance lost?}
\label{sec:decay}

A parallel drafter proposes a whole block in one pass, so a longer block adds little drafting cost.
We investigate how acceptance length varies as a function of block size. 
Figure~\ref{fig:proplen8b} shows acceptance length when the drafters, all trained at $K=16$, propose blocks of $K=2$ to $16$ tokens on Qwen3-8B (\citealt{chen_dflash_2026} show that the penalty of such a training-inference mismatch in $K$ is ${\sim}0.15$ tokens), and Figure~\ref{fig:decay8b} shows the conditional acceptance rate at each slot of a 16-token block (the acceptance rate at a given slot, assuming that all tokens before that position were accepted).

Acceptance length generally grows with $K$ up to the trained block length, with $\gamma$ growing more slowly for chat than for math and code (note the different y-axes).
The conditional acceptance curves show where the drafters differ.
One-step \ours{} matches or exceeds DFlash at the first slot in every domain; however, the conditional acceptance decays more rapidly for \ours{} at 1 NFE than for DFlash.
This indicates that the later slots in the block are less likely to be accepted.
For \ours{} with two steps, the rate at slot 16 is higher than for DFlash across all domains.
Because a gain at an early slot counts toward every block that reaches it, while a loss at a late slot affects only the few blocks that get that far, one-step \ours{} still accepts more than DFlash overall (Table~\ref{tab:acceptance}), but it gains less than DFlash from its last slots.

Notably, two-step \ours{}'s conditional acceptance does not decay within the block for any domain.
The conditional acceptance curve stays flat, or even rises, matching what is observed in models that apply sequential postprocessing to re-capture inter-token dependencies \citep{cheng_dspark_2026}.
This indicates that it may gain more than DFlash or one-step \ours{} from blocks even longer than 16 tokens, but testing this requires drafters trained at larger $K$.
Why one-step drafts, which come from a joint model, fall below a factorized drafter at late positions in the block is the question we turn to next.

\subsection{Does \ours{} capture joint structure?}
\label{sec:shuffle}

Whether a draft is accepted depends on two properties of the drafter's distribution over blocks: its per-slot marginals, i.e., how often each slot's token matches the target on its own, and the dependence between slots, i.e., whether the tokens in a block are consistent with each other.
DFlash has only marginals: it predicts each slot independently and, at $T=0$, drafts the most likely token of each.
\ours{} drafts the whole block from one noise draw, which can make the block consistent, but each slot's token then depends on the noise, so it is not necessarily the most likely token under that slot's marginal.
The conditional acceptance rates in Figure~\ref{fig:decay8b} reflect both properties and cannot separate them.
We separate them with a shuffle test: we compare drafts whose slots all come from one shared noise draw (\textit{shared}) with drafts using a different, independent noise draw per slot (\textit{shuffled}).
Shuffling keeps each slot's marginal but removes the dependence between slots, so $\gamma_{\mathrm{shuffled}}$ measures the marginals alone and $\Delta = \gamma_{\mathrm{shared}} - \gamma_{\mathrm{shuffled}}$ is the acceptance gain due to the draft's joint structure.
A drafter whose proposal does not depend on the noise, such as DFlash, has $\Delta = 0$ by construction, whereas a drafter that exploits joint structure has $\Delta>0$.
We show the results of the shuffle test in Table \ref{tab:shuffle8b}.

\paragraph{Integration creates joint structure.}
Joint structure is modelled along a flow's trajectory.
Indeed, when performing only a single evaluation of the flow model, we find that there is no difference in acceptance length between the shared noise configuration and the shuffled noise configuration, indicating that the flow captures no joint structure at 1 NFE.
However, the same flow, when integrated for 16 steps, has $\Delta>0$, and indeed, can attribute almost an entire token's worth of acceptance to joint structure.
This indicates that it is integration of the flow that resolves each noise draw into one consistent block.
It is this integration that we try to distill into a few-step flow map in \ours{}.

\paragraph{Two steps capture most of it.}
At 1 NFE, \ours{} retains only a small fraction of the $\Delta$ in the 16-step teacher, while at 2-NFE, \ours{} recovers most of the $\Delta$ in the 16-step teacher.
This indicates that a single jump of the flow map from pure noise underutilizes the noise, and produces a draft that, while not totally factorized, does not fully capture all inter-token dependencies---consistent with its falling conditional acceptance in Figure~\ref{fig:decay8b}.
In contrast, the 2 NFE flow map heavily exploits inter-token dependencies, which explains its flat or increasing conditional acceptance along the block (Figure~\ref{fig:decay8b}).

\begin{table*}[!t]
  \centering
  \small
  \setlength{\tabcolsep}{5pt}
  \begin{tabular}{@{}lccccccccc@{}}
    \toprule
    & \multicolumn{3}{c}{Math} & \multicolumn{3}{c}{Code} & \multicolumn{3}{c}{Chat} \\
    \cmidrule(lr){2-4} \cmidrule(lr){5-7} \cmidrule(l){8-10}
    Drafter & Shared & Shuffled & $\Delta$ & Shared & Shuffled & $\Delta$ & Shared & Shuffled & $\Delta$ \\
    \midrule
    DFlash        & 7.29 & 7.29 & 0.00 & 6.07 & 6.07 & 0.00 & 2.92 & 2.92 & 0.00 \\
    \addlinespace
    Flow (1 step) & 6.56 & 6.56 & 0.00 & 5.46 & 5.46 & 0.00 & 2.88 & 2.88 & 0.00 \\
    Flow (16 steps)         & 7.85 & 7.03 & 0.82 & 6.70 & 6.11 & 0.59 & 3.32 & 3.03 & 0.29 \\
    \addlinespace
    \ours{} (1 step) & 7.05 & 6.73 & 0.32 & 5.92 & 5.78 & 0.13 & 3.04 & 2.96 & 0.08 \\
    \ours{} (2 steps) & 7.70 & 6.92 & 0.78 & 6.57 & 6.11 & 0.47 & 3.27 & 3.03 & 0.25 \\
    \bottomrule
  \end{tabular}
  \caption{Shuffle test on Qwen3-8B at $T=0$: accepted draft tokens per block, for a draft with shared noise across slots (shared) and for a draft with independent per-slot noise (shuffled). $\Delta$ is the difference between the two, with a larger value indicating a greater acceptance due to joint structure. Results for Qwen3-4B are in Table \ref{tab:shuffle4b}.}
  \label{tab:shuffle8b}
\end{table*}

\subsection{What limits one-step drafting?}
\label{sec:limits}

At both 1 NFE and 2 NFE, \ours{} trails the flow teacher at 16 NFE in terms of acceptance length under the shared setting, which is the usual inference setting.
A better distillation strategy would reduce this gap and make the 1 NFE flow map competitive with the many-NFE flow teacher.
Further, \ours{} also trails the teacher in the shuffled setting, indicating that the per-position marginals are also less well estimated.

While in theory flow map distillation enables few-step flow maps to capture inter-token dependencies, these results indicate that there is still a gap between few-step flow map models and one-step models.
Indeed, some recent works on flow map language models find that while performance of such models is strong in the few-step regime, one-step models still suffer from high generative perplexity or collapsed entropy \citep[e.g.,][]{potaptchik_discrete_2026, tang2026discretebeckmanntransportmodels}.
Our results also indicate potential for future work to improve, both on the theoretical and empirical sides, the distillation strategies for discrete flow maps, particularly for inference at 1 NFE.

\section{Related Work}

\paragraph{Speculative decoding.}
Speculative decoding drafts tokens with a cheap model and verifies them with the target in one parallel pass, while keeping the target's output distribution~\citep{leviathan_fast_2023,chen_accelerating_2023}.
Most drafters are autoregressive. EAGLE-3~\citep{li_eagle-3_2025} conditions a lightweight drafter on the target's hidden states and reaches high acceptance, but it drafts one token per forward pass, so its drafting cost grows with block length.

\paragraph{Diffusion drafters.}
Discrete diffusion drafters instead propose a whole block in parallel.
DiffuSpec~\citep{li_diffuspec_2026} and SpecDiff-2~\citep{sandler_specdiff-2_2026} repurpose large pretrained diffusion LMs, adding causal path search and verifier calibration respectively, but their drafter size limits practical speedups.
DFlash~\citep{chen_dflash_2026} trains a lightweight block diffusion drafter on the target's hidden states and drafts a block in a single forward pass.
Recent work adds draft trees on top of DFlash to accelerate sampling~\citep{ringel_accelerating_2026}.
Since a one-pass discrete drafter like DFlash predicts each position independently (Eq.~\ref{eq:factorized}), follow-up works attempt to recover joint structure from this factorized output through autoregressive postprocessing~\citep{cheng_dspark_2026, huang2026dominodecouplingcausalmodeling, zheng2026delsspecdecoupledlongshortcontexts} or low-rank latent mixtures~\citep{karimi_dblast_2026}.
These methods add search or sequential computation on top of the parallel draft, or restrict the joint to a low-rank family.

\paragraph{Flow maps for discrete data.}
Alongside Flow Map Language Models~\citep{lee_flow_2026}, Discrete Flow Maps~\citep{potaptchik_discrete_2026} and Categorical Flow Maps~\citep{roos_categorical_2026} learn flow maps that respect the geometry of the probability simplex, achieving strong few-step generation of text and other categorical data.
Instead of standalone generation, we focus on drafting for a target model.

\paragraph{Flow models as drafters.}
To our knowledge, \ours{} is the first speculative decoding drafter built on a continuous flow.
Rather than correcting a factorized proposal after drafting, it changes the drafter's generative model, so that dependencies between slots can be represented within the drafting pass itself.
\ours{} keeps DFlash's conditioning on target hidden states and its frozen embedding and LM head, and replaces its discrete diffusion model with a continuous flow.

\section*{Limitations}

We evaluate two target models from one family, Qwen3-4B and Qwen3-8B; other model families and larger targets are left to future work.
Because of compute limits, all drafters, including the baselines, are trained on a 500K-sample subset of Open-PerfectBlend, so absolute acceptance lengths may differ from those of drafters trained on the full data.
Speedups are measured at concurrency 1 on a single GPU, and batched serving is not evaluated.
We draft blocks of 16 tokens and compare against EAGLE-3 at the same 16-token verification budget.
Our baselines are EAGLE-3 and DFlash; we do not evaluate follow-ups that build on DFlash's draft, such as draft trees~\citep{ringel_accelerating_2026}, autoregressive postprocessing~\citep{cheng_dspark_2026}, or low-rank latent mixtures~\citep{karimi_dblast_2026}, although \ours{} is also compatible with draft trees (Appendix~\ref{sec:appendix-multisample}).
Our benchmarks cover English math, code, and chat at temperatures 0 and 1.
It is also possible that our models are undertrained;
for instance, \citealt{lee_flow_2026} train their flow language model over 1M steps and distill over 100K steps, while we train and distill only for 39K and 23K steps, respectively, due to compute constraints.

\section*{Ethical Considerations}

\paragraph{Risks.}
\ours{} is a drafter for speculative decoding, which produces a system that preserves the sampling distribution of the target model.
The risks of the target model, such as biased, harmful, or factually wrong outputs, therefore carry over. As with any inference speedup, lowering the cost of generation also lowers the cost of misuse at scale.

\paragraph{Artifacts and licenses.}
We use only publicly available models, data, and software, and cite their creators (\S\ref{sec:setup}).
Qwen3-4B and Qwen3-8B and SGLang are released under the Apache~2.0 license, Open-PerfectBlend under Apache~2.0, and DeepSpec under MIT.
The evaluation benchmarks are used for evaluation only, consistent with their intended use and licenses (Appendix~\ref{sec:appendix-impl}).

\paragraph{Data.}
We do not collect new data or use human annotators.
Open-PerfectBlend is a public mixture of existing instruction-tuning datasets, mostly in English, covering math, code, and chat.

\paragraph{Use of AI assistants.}
We used AI assistants to help write experiment code and to edit the text of this paper. All ideas, experiments, and results are the authors' own, and the authors reviewed and verified all AI-assisted code and text.

\section*{Acknowledgments}

J.S. is supported by the Natural Sciences and Engineering Research Council of Canada (NSERC), funding reference number 587652. 
We would like to acknowledge the support of the Simons Foundation,  Schmidt Sciences, and Coefficient Giving.
This work was supported in part by the AI2050 program at Schmidt Sciences (Grant G-25-70028). 
Additionally, computations were run at facilities supported by the Scientific Computing Core at the Flatiron Institute.
The Flatiron Institute is a division of the Simons Foundation. 

\bibliography{custom}

@misc{chen_dflash_2026,
	title = {{DFlash}: {Block} {Diffusion} for {Flash} {Speculative} {Decoding}},
	shorttitle = {{DFlash}},
	url = {http://arxiv.org/abs/2602.06036},
	doi = {10.48550/arXiv.2602.06036},
	urldate = {2026-09-25},
	publisher = {arXiv},
	author = {Chen, Jian and Liang, Yesheng and Liu, Zhijian},
	month = may,
	year = {2026},
	note = {arXiv:2602.06036 [cs.CL]},
}

@inproceedings{li_diffuspec_2026,
	address = {San Diego, California, United States},
	title = {{DiffuSpec}: {Unlocking} {Diffusion} {Language} {Models} for {Speculative} {Decoding}},
	isbn = {979-8-89176-395-1},
	shorttitle = {{DiffuSpec}},
	url = {https://aclanthology.org/2026.findings-acl.1048/},
	doi = {10.18653/v1/2026.findings-acl.1048},
	urldate = {2026-09-25},
	booktitle = {Findings of the {Association} for {Computational} {Linguistics}: {ACL} 2026},
	publisher = {Association for Computational Linguistics},
	author = {Li, Guanghao and Fu, Zhihui and Fang, Min and Zhao, Qibin and Tang, Ming and Yuan, Chun and Wang, Jun},
	editor = {Liakata, Maria and Moreira, Viviane P. and Zhang, Jiajun and Jurgens, David},
	month = jul,
	year = {2026},
	pages = {20896--20910},
}

@article{sandler_specdiff-2_2026,
	title = {{SpecDiff}-2: {Scaling} {Diffusion} {Drafter} {Alignment} {For} {Faster} {Speculative} {Decoding}},
	volume = {8},
	shorttitle = {{SpecDiff}-2},
	url = {https://proceedings.mlsys.org/paper_files/paper/2026/hash/041dad5ed2191b44ba3ed0e00cdc3187-Abstract-Conference.html},
	language = {en},
	urldate = {2026-09-25},
	journal = {Proceedings of Machine Learning and Systems},
	author = {Sandler, Jameson and Christopher, Jacob and Hartvigsen, Tom and Fioretto, Ferdinando},
	month = may,
	year = {2026},
	pages = {1128--1147},
}

@misc{ringel_accelerating_2026,
	title = {Accelerating {Speculative} {Decoding} with {Block} {Diffusion} {Draft} {Trees}},
	url = {http://arxiv.org/abs/2604.12989},
	doi = {10.48550/arXiv.2604.12989},
	urldate = {2026-09-25},
	publisher = {arXiv},
	author = {Ringel, Liran and Romano, Yaniv},
	month = apr,
	year = {2026},
	note = {arXiv:2604.12989 [cs.CL]},
}

@misc{lee_flow_2026,
	title = {Flow {Map} {Language} {Models}: {One}-step {Language} {Modeling} via {Continuous} {Denoising}},
	shorttitle = {Flow {Map} {Language} {Models}},
	url = {http://arxiv.org/abs/2602.16813},
	doi = {10.48550/arXiv.2602.16813},
	urldate = {2026-09-25},
	publisher = {arXiv},
	author = {Lee, Chanhyuk and Yoo, Jaehoon and Agarwal, Manan and Shah, Sheel and Huang, Jerry and Raghunathan, Aditi and Hong, Seunghoon and Boffi, Nicholas M. and Kim, Jinwoo},
	month = may,
	year = {2026},
	note = {arXiv:2602.16813 [cs.CL]},
}

@misc{potaptchik_discrete_2026,
	title = {Discrete {Flow} {Maps}},
	url = {https://arxiv.org/abs/2604.09784v2},
	language = {en},
	urldate = {2026-09-29},
	journal = {arXiv.org},
	author = {Potaptchik, Peter and Yim, Jason and Saravanan, Adhi and Holderrieth, Peter and Vanden-Eijnden, Eric and Albergo, Michael S.},
	month = apr,
	year = {2026},
}

@misc{roos_categorical_2026,
	title = {Categorical {Flow} {Maps}},
	copyright = {Creative Commons Attribution 4.0 International},
	url = {https://arxiv.org/abs/2602.12233},
	doi = {10.48550/ARXIV.2602.12233},
	language = {en},
	urldate = {2026-09-29},
	publisher = {arXiv},
	author = {Roos, Daan and Davis, Oscar and Eijkelboom, Floor and Bronstein, Michael and Welling, Max and Ceylan, İsmail İlkan and Ambrogioni, Luca and van de Meent, Jan-Willem},
	year = {2026},
	note = {Version Number: 1},
}

@misc{li_eagle-3_2025,
	title = {{EAGLE}-3: {Scaling} up {Inference} {Acceleration} of {Large} {Language} {Models} via {Training}-{Time} {Test}},
	shorttitle = {{EAGLE}-3},
	url = {https://arxiv.org/abs/2503.01840v1},
	language = {en},
	urldate = {2026-09-29},
	journal = {arXiv.org},
	author = {Li, Yuhui and Wei, Fangyun and Zhang, Chao and Zhang, Hongyang},
	month = mar,
	year = {2025},
}

@misc{cheng_dspark_2026,
	title = {{DSpark}: {Confidence}-{Scheduled} {Speculative} {Decoding} with {Semi}-{Autoregressive} {Generation}},
	shorttitle = {{DSpark}},
	url = {http://arxiv.org/abs/2607.05147},
	doi = {10.48550/arXiv.2607.05147},
	urldate = {2026-09-29},
	publisher = {arXiv},
	author = {Cheng, Xin and Yu, Xingkai and Shao, Chenze and Li, Jiashi and Xiong, Yunfan and Qian, Yi and Zhu, Jiaqi and Ma, Shirong and Zhang, Xiaokang and Ye, Jiasheng and Chen, Qinyu and Deng, Chengqi and Yu, Jiping and Dai, Damai and Zhang, Zhengyan and Wei, Yixuan and Tan, Yixuan and Yang, Wenkai and Xu, Runxin and Wu, Yu and Xu, Zhean and Wang, Xuanyu and Chen, Muyang and Tian, Rui and Bi, Xiao and Hao, Zhewen and Chen, Shaoyuan and Cao, Huanqi and Zhang, Wentao and Xu, Anyi and Zhang, Huishuai and Zhao, Dongyan and Liang, Wenfeng},
	month = jul,
	year = {2026},
	note = {arXiv:2607.05147 [cs.AI]},
}

@misc{karimi_dblast_2026,
	title = {{DBLAST}: {Dependent} {Block} {Drafting} for {Stochastic} {Speculative} {Decoding}},
	shorttitle = {{DBLAST}},
	url = {https://arxiv.org/abs/2608.05448v1},
	language = {en},
	urldate = {2026-09-29},
	journal = {arXiv.org},
	author = {Karimi, Amirmohammad and Gao, Chao and Hassanpour, Negar},
	month = aug,
	year = {2026},
}

@misc{leviathan_fast_2023,
      title={Fast Inference from Transformers via Speculative Decoding}, 
      author={Yaniv Leviathan and Matan Kalman and Yossi Matias},
      year={2023},
      eprint={2211.17192},
      archivePrefix={arXiv},
      primaryClass={cs.LG},
      url={https://arxiv.org/abs/2211.17192}, 
}

@misc{chen_accelerating_2023,
      title={Accelerating Large Language Model Decoding with Speculative Sampling}, 
      author={Charlie Chen and Sebastian Borgeaud and Geoffrey Irving and Jean-Baptiste Lespiau and Laurent Sifre and John Jumper},
      year={2023},
      eprint={2302.01318},
      archivePrefix={arXiv},
      primaryClass={cs.CL},
      url={https://arxiv.org/abs/2302.01318}, 
}

@misc{lipman_flow_2023,
      title={Flow Matching for Generative Modeling}, 
      author={Yaron Lipman and Ricky T. Q. Chen and Heli Ben-Hamu and Maximilian Nickel and Matt Le},
      year={2023},
      eprint={2210.02747},
      archivePrefix={arXiv},
      primaryClass={cs.LG},
      url={https://arxiv.org/abs/2210.02747}, 
}

@misc{liu2022rf,
      title={Flow Straight and Fast: Learning to Generate and Transfer Data with Rectified Flow}, 
      author={Xingchao Liu and Chengyue Gong and Qiang Liu},
      year={2022},
      eprint={2209.03003},
      archivePrefix={arXiv},
      primaryClass={cs.LG},
      url={https://arxiv.org/abs/2209.03003}, 
}

@misc{albergo_stochastic_2025,
      title={Stochastic Interpolants: A Unifying Framework for Flows and Diffusions}, 
      author={Michael S. Albergo and Nicholas M. Boffi and Eric Vanden-Eijnden},
      year={2025},
      eprint={2303.08797},
      archivePrefix={arXiv},
      primaryClass={cs.LG},
      url={https://arxiv.org/abs/2303.08797}, 
}

@misc{boffi2025flowmap,
      title={Flow map matching with stochastic interpolants: A mathematical framework for consistency models}, 
      author={Nicholas M. Boffi and Michael S. Albergo and Eric Vanden-Eijnden},
      year={2025},
      eprint={2406.07507},
      archivePrefix={arXiv},
      primaryClass={cs.LG},
      url={https://arxiv.org/abs/2406.07507}, 
}

@misc{boffi2025consistency,
      title={How to build a consistency model: Learning flow maps via self-distillation}, 
      author={Nicholas M. Boffi and Michael S. Albergo and Eric Vanden-Eijnden},
      year={2025},
      eprint={2505.18825},
      archivePrefix={arXiv},
      primaryClass={cs.LG},
      url={https://arxiv.org/abs/2505.18825}, 
}

@misc{peebles_scalable_2023,
      title={Scalable Diffusion Models with Transformers}, 
      author={William Peebles and Saining Xie},
      year={2023},
      eprint={2212.09748},
      archivePrefix={arXiv},
      primaryClass={cs.CV},
      url={https://arxiv.org/abs/2212.09748}, 
}

@misc{li2026basics,
      title={Back to Basics: Let Denoising Generative Models Denoise}, 
      author={Tianhong Li and Kaiming He},
      year={2026},
      eprint={2511.13720},
      archivePrefix={arXiv},
      primaryClass={cs.CV},
      url={https://arxiv.org/abs/2511.13720}, 
}

@misc{zheng2026delsspecdecoupledlongshortcontexts,
      title={DeLS-Spec: Decoupled Long-Short Contexts for Parallel Speculative Drafting}, 
      author={Hong-Kai Zheng and Piji Li},
      year={2026},
      eprint={2607.07409},
      archivePrefix={arXiv},
      primaryClass={cs.CL},
      url={https://arxiv.org/abs/2607.07409}, 
}

@misc{huang2026dominodecouplingcausalmodeling,
      title={Domino: Decoupling Causal Modeling from Autoregressive Drafting in Speculative Decoding}, 
      author={Jianuo Huang and Yaojie Zhang and Qituan Zhang and Hao Lin and Hanlin Xu and Linfeng Zhang},
      year={2026},
      eprint={2605.29707},
      archivePrefix={arXiv},
      primaryClass={cs.CL},
      url={https://arxiv.org/abs/2605.29707}, 
}

@misc{yang2025qwen3technicalreport,
      title={Qwen3 Technical Report}, 
      author={An Yang and Anfeng Li and Baosong Yang and Beichen Zhang and Binyuan Hui and Bo Zheng and Bowen Yu and Chang Gao and Chengen Huang and Chenxu Lv and Chujie Zheng and Dayiheng Liu and Fan Zhou and Fei Huang and Feng Hu and Hao Ge and Haoran Wei and Huan Lin and Jialong Tang and Jian Yang and Jianhong Tu and Jianwei Zhang and Jianxin Yang and Jiaxi Yang and Jing Zhou and Jingren Zhou and Junyang Lin and Kai Dang and Keqin Bao and Kexin Yang and Le Yu and Lianghao Deng and Mei Li and Mingfeng Xue and Mingze Li and Pei Zhang and Peng Wang and Qin Zhu and Rui Men and Ruize Gao and Shixuan Liu and Shuang Luo and Tianhao Li and Tianyi Tang and Wenbiao Yin and Xingzhang Ren and Xinyu Wang and Xinyu Zhang and Xuancheng Ren and Yang Fan and Yang Su and Yichang Zhang and Yinger Zhang and Yu Wan and Yuqiong Liu and Zekun Wang and Zeyu Cui and Zhenru Zhang and Zhipeng Zhou and Zihan Qiu},
      year={2025},
      eprint={2505.09388},
      archivePrefix={arXiv},
      primaryClass={cs.CL},
      url={https://arxiv.org/abs/2505.09388}, 
}

@misc{xu2024perfectblendredefiningrlhf,
      title={The Perfect Blend: Redefining RLHF with Mixture of Judges}, 
      author={Tengyu Xu and Eryk Helenowski and Karthik Abinav Sankararaman and Di Jin and Kaiyan Peng and Eric Han and Shaoliang Nie and Chen Zhu and Hejia Zhang and Wenxuan Zhou and Zhouhao Zeng and Yun He and Karishma Mandyam and Arya Talabzadeh and Madian Khabsa and Gabriel Cohen and Yuandong Tian and Hao Ma and Sinong Wang and Han Fang},
      year={2024},
      eprint={2409.20370},
      archivePrefix={arXiv},
      primaryClass={cs.LG},
      url={https://arxiv.org/abs/2409.20370}, 
}

@misc{loshchilov2019decoupledweightdecayregularization,
      title={Decoupled Weight Decay Regularization}, 
      author={Ilya Loshchilov and Frank Hutter},
      year={2019},
      eprint={1711.05101},
      archivePrefix={arXiv},
      primaryClass={cs.LG},
      url={https://arxiv.org/abs/1711.05101}, 
}

@misc{cobbe2021trainingverifierssolvemath,
      title={Training Verifiers to Solve Math Word Problems}, 
      author={Karl Cobbe and Vineet Kosaraju and Mohammad Bavarian and Mark Chen and Heewoo Jun and Lukasz Kaiser and Matthias Plappert and Jerry Tworek and Jacob Hilton and Reiichiro Nakano and Christopher Hesse and John Schulman},
      year={2021},
      eprint={2110.14168},
      archivePrefix={arXiv},
      primaryClass={cs.LG},
      url={https://arxiv.org/abs/2110.14168}, 
}

@misc{hendrycks2021measuringmathematicalproblemsolving,
      title={Measuring Mathematical Problem Solving With the MATH Dataset}, 
      author={Dan Hendrycks and Collin Burns and Saurav Kadavath and Akul Arora and Steven Basart and Eric Tang and Dawn Song and Jacob Steinhardt},
      year={2021},
      eprint={2103.03874},
      archivePrefix={arXiv},
      primaryClass={cs.LG},
      url={https://arxiv.org/abs/2103.03874}, 
}

@misc{aime25,
      title={American Invitational Mathematics Examination (AIME) 2025}, 
      author={Zhang, Yifan and Math-AI, Team},
      year={2025},
}

@misc{austin2021programsynthesislargelanguage,
      title={Program Synthesis with Large Language Models}, 
      author={Jacob Austin and Augustus Odena and Maxwell Nye and Maarten Bosma and Henryk Michalewski and David Dohan and Ellen Jiang and Carrie Cai and Michael Terry and Quoc Le and Charles Sutton},
      year={2021},
      eprint={2108.07732},
      archivePrefix={arXiv},
      primaryClass={cs.PL},
      url={https://arxiv.org/abs/2108.07732}, 
}

@misc{chen2021evaluating,
      title={Evaluating Large Language Models Trained on Code},
      author={Mark Chen and Jerry Tworek and Heewoo Jun and Qiming Yuan and Henrique Ponde de Oliveira Pinto and Jared Kaplan and Harri Edwards and Yuri Burda and Nicholas Joseph and Greg Brockman and Alex Ray and Raul Puri and Gretchen Krueger and Michael Petrov and Heidy Khlaaf and Girish Sastry and Pamela Mishkin and Brooke Chan and Scott Gray and Nick Ryder and Mikhail Pavlov and Alethea Power and Lukasz Kaiser and Mohammad Bavarian and Clemens Winter and Philippe Tillet and Felipe Petroski Such and Dave Cummings and Matthias Plappert and Fotios Chantzis and Elizabeth Barnes and Ariel Herbert-Voss and William Hebgen Guss and Alex Nichol and Alex Paino and Nikolas Tezak and Jie Tang and Igor Babuschkin and Suchir Balaji and Shantanu Jain and William Saunders and Christopher Hesse and Andrew N. Carr and Jan Leike and Josh Achiam and Vedant Misra and Evan Morikawa and Alec Radford and Matthew Knight and Miles Brundage and Mira Murati and Katie Mayer and Peter Welinder and Bob McGrew and Dario Amodei and Sam McCandlish and Ilya Sutskever and Wojciech Zaremba},
      year={2021},
      eprint={2107.03374},
      archivePrefix={arXiv},
      primaryClass={cs.LG}
}

@misc{jain2024livecodebenchholisticcontaminationfree,
      title={LiveCodeBench: Holistic and Contamination Free Evaluation of Large Language Models for Code}, 
      author={Naman Jain and King Han and Alex Gu and Wen-Ding Li and Fanjia Yan and Tianjun Zhang and Sida Wang and Armando Solar-Lezama and Koushik Sen and Ion Stoica},
      year={2024},
      eprint={2403.07974},
      archivePrefix={arXiv},
      primaryClass={cs.SE},
      url={https://arxiv.org/abs/2403.07974}, 
}

@inproceedings{Bai_2024,
   title={MT-Bench-101: A Fine-Grained Benchmark for Evaluating Large Language Models in Multi-Turn Dialogues},
   url={http://dx.doi.org/10.18653/v1/2024.acl-long.401},
   DOI={10.18653/v1/2024.acl-long.401},
   booktitle={Proceedings of the 62nd Annual Meeting of the Association for Computational Linguistics (Volume 1: Long Papers)},
   publisher={Association for Computational Linguistics},
   author={Bai, Ge and Liu, Jie and Bu, Xingyuan and He, Yancheng and Liu, Jiaheng and Zhou, Zhanhui and Lin, Zhuoran and Su, Wenbo and Ge, Tiezheng and Zheng, Bo and Ouyang, Wanli},
   year={2024},
   pages={7421–7454} }

@misc{alpaca_eval,
  author = {Xuechen Li and Tianyi Zhang and Yann Dubois and Rohan Taori and Ishaan Gulrajani and Carlos Guestrin and Percy Liang and Tatsunori B. Hashimoto },
  title = {AlpacaEval: An Automatic Evaluator of Instruction-following Models},
  year = {2023},
  month = {5},
  publisher = {GitHub},
  journal = {GitHub repository},
  howpublished = {\url{https://github.com/tatsu-lab/alpaca_eval}}
}

@misc{li2024crowdsourceddatahighqualitybenchmarks,
      title={From Crowdsourced Data to High-Quality Benchmarks: Arena-Hard and BenchBuilder Pipeline}, 
      author={Tianle Li and Wei-Lin Chiang and Evan Frick and Lisa Dunlap and Tianhao Wu and Banghua Zhu and Joseph E. Gonzalez and Ion Stoica},
      year={2024},
      eprint={2406.11939},
      archivePrefix={arXiv},
      primaryClass={cs.LG},
      url={https://arxiv.org/abs/2406.11939}, 
}

@misc{tang2026discretebeckmanntransportmodels,
      title={Discrete Beckmann Transport Models for One-Step Language Modeling and Reasoning}, 
      author={Sophia Tang and Shiyi Wang},
      year={2026},
      eprint={2609.15903},
      archivePrefix={arXiv},
      primaryClass={cs.LG},
      url={https://arxiv.org/abs/2609.15903}, 
}

@misc{gu2018nonautoregressiveneuralmachinetranslation,
      title={Non-Autoregressive Neural Machine Translation}, 
      author={Jiatao Gu and James Bradbury and Caiming Xiong and Victor O. K. Li and Richard Socher},
      year={2018},
      eprint={1711.02281},
      archivePrefix={arXiv},
      primaryClass={cs.CL},
      url={https://arxiv.org/abs/1711.02281}, 
}

\appendix

\section{Input featurization}

A frozen input projection constrains how the state can enter the network.
The natural choice, the linear read-out $x^k E$ of the state at each slot, sums the embeddings of all tokens, each weighted by its noisy coordinate.
The clean token contributes only one term and every other token contributes noise, so the signal is drowned out.
For the flow model, we therefore embed the posterior over the clean token given the state, which provides a better signal.
Under a uniform prior, the posterior over classes given a Gaussian-noised one-hot is
\begin{equation}
  \pi_t(x^k) = \mathrm{softmax}\left(\frac{t\,x^k}{(1-t)^2}\right).
  \label{eq:posterior}
\end{equation}
The network input $h^k = \mathrm{RMSNorm}(\pi_t(x^k)\,E)$ is then dominated by the embeddings of the few tokens consistent with the state.
We find, in small-scale tests, that this improves the performance of the flow teacher model as compared to linear featurization. 

At $s = 0$, however, this posterior input $\pi_0(x^k)$ is uniform for every $x^k$.
For a flow map model that will be evaluated with 1 NFE (i.e., for $(s, t) = (0, 1)$), we require that the model has different random noise to anchor its joint prediction.
That is, in the case where we only feed in the posterior featurization to the flow map model and evaluate it at 1 NFE, the flow map model will learn the per-slot marginals and revert back to a deterministic factorized drafter. 
The student therefore also reads the raw state through a linear path,
\begin{equation}
  h^k = \pi_s(x^k)\,E + (t-s)\,x^k E,
\end{equation}
where the $t-s$ term ensures that the student matches the teacher on the diagonal, and each term is normalized with RMSNorm.

\section{Fast multi-step drafting}
\label{sec:appendix-fast}
Drafting with more than one jump embeds a vocabulary-sized state at every slot: $x_0 E$ at the first jump, and $\pi_u(x_u)\,E$ and $x_u E$ at the second, each a full pass over $E$.
With two jumps at times $0 < u < 1$, the state entering the second jump is $x_u = (1-u)\,x_0 + u\,\hat\delta$, where $\hat\delta = \delta^\phi_{0,u}(x_0; c)$, so its posterior is $\pi_u(x_u^k) \propto \exp(\alpha x_0^k + \alpha^2 \hat\delta^k)$ with $\alpha = u/(1-u)$.
Since $\hat\delta^k$ is concentrated on a few tokens, we split each token's weight into a noise-only part and a correction on the top tokens $T^k$ of $\hat\delta^k$:
\begin{equation}
  \begin{gathered}
    \pi_u(x_u^k)\,E \approx \frac{W(x_0^k) + \sum_{v \in T^k} r_v E_v}{S(x_0^k) + \sum_{v \in T^k} r_v}, \\
    W(z) = \sum_{v \in \mathcal{V}} e^{\alpha z_v} E_v, \qquad S(z) = \sum_{v \in \mathcal{V}} e^{\alpha z_v},
  \end{gathered}
  \label{eq:fastpath}
\end{equation}
where $r_v = e^{\alpha x^k_{0,v}}\big(e^{\alpha^2\hat\delta^k_v} - 1\big)$; likewise $x_u^k E \approx (1-u)\,x_0^k E + u \sum_{v \in T^k} \hat\delta^k_v E_v$.
The parts over the whole vocabulary, $x_0 E$, $W$, and $S$, depend only on the noise and on $\alpha$, which we can precompute for a bank of 8{,}192 noise draws.
At drafting time each slot selects a random bank entry, so $x_0 E$, $W$, and $S$ are already precomputed, and we only require computing the correction.
The correction needs the noise on $T^k$, which we redraw; this perturbs only $|T^k|$ of the $|\mathcal{V}|$ terms in $W$ and $S$.
With $|T^k| = 16$, the second jump reads 16 rows of $E$ per slot instead of all of them, reducing the added cost of the second jump over one-jump drafting by about 70\% on Qwen3-8B, while acceptance stays essentially unchanged as compared to the two-step draft without this optimization.

\section{Speedup and acceptance length on Qwen3-4B}
\label{sec:appendix-4b}

Table~\ref{tab:speedup-4b} gives speedup and acceptance length on Qwen3-4B.
These results are analogous to the ones presented in Figure~\ref{fig:overview}(c) and Table \ref{tab:acceptance}.

\begin{table*}[!t]
  \centering
  \small
  \setlength{\tabcolsep}{3.5pt}
  \begin{tabular}{@{}llccccccccc@{}}
    \toprule
    \multirow{2}{*}{Temp.} & \multirow{2}{*}{Method} & \multicolumn{3}{c}{Math} & \multicolumn{3}{c}{Code} & \multicolumn{3}{c}{Chat} \\
    \cmidrule(lr){3-5} \cmidrule(lr){6-8} \cmidrule(l){9-11}
    & & GSM8K & MATH & AIME25 & MBPP & HumanEval & LCB & MT-Bench & Alpaca & Arena-Hard \\
    \midrule
    \multicolumn{11}{c}{\textit{Speedup ($\times$)}} \\
    \midrule
    \multirow{4}{*}{$T=0$}
      & EAGLE-3 (16) & 2.66 & 2.64 & 2.47 & 2.20 & 2.33 & 2.44 & 1.49 & 1.37 & 1.64 \\
      & DFlash & \textbf{5.18} & \textbf{5.23} & \textbf{4.56} & \textbf{3.72} & \underline{4.00} & \underline{4.27} & \underline{2.43} & \underline{2.28} & \underline{2.58} \\
      & \cellcolor{oursrow}\ours{} (1 step) & \cellcolor{oursrow}\underline{5.14} & \cellcolor{oursrow}\underline{5.02} & \cellcolor{oursrow}\underline{4.47} & \cellcolor{oursrow}\underline{3.71} & \cellcolor{oursrow}\textbf{4.04} & \cellcolor{oursrow}\textbf{4.28} & \cellcolor{oursrow}\textbf{2.46} & \cellcolor{oursrow}\textbf{2.28} & \cellcolor{oursrow}\textbf{2.66} \\
      & \cellcolor{oursrow}\ours{} (2 steps) & \cellcolor{oursrow}4.75 & \cellcolor{oursrow}4.57 & \cellcolor{oursrow}4.00 & \cellcolor{oursrow}3.42 & \cellcolor{oursrow}3.76 & \cellcolor{oursrow}3.87 & \cellcolor{oursrow}2.16 & \cellcolor{oursrow}2.00 & \cellcolor{oursrow}2.34 \\
    \addlinespace
    \multirow{4}{*}{$T=1$}
      & EAGLE-3 (16) & 2.57 & 2.31 & 1.94 & 2.04 & 2.21 & 2.16 & 1.37 & 1.29 & 1.41 \\
      & DFlash & \textbf{4.67} & \textbf{3.96} & \textbf{3.29} & \underline{3.26} & \underline{3.67} & \textbf{3.25} & \underline{2.16} & \textbf{2.15} & \underline{2.02} \\
      & \cellcolor{oursrow}\ours{} (1 step) & \cellcolor{oursrow}\underline{4.59} & \cellcolor{oursrow}\underline{3.81} & \cellcolor{oursrow}\underline{3.12} & \cellcolor{oursrow}\textbf{3.26} & \cellcolor{oursrow}\textbf{3.69} & \cellcolor{oursrow}\underline{3.24} & \cellcolor{oursrow}\textbf{2.25} & \cellcolor{oursrow}\underline{2.04} & \cellcolor{oursrow}\textbf{2.17} \\
      & \cellcolor{oursrow}\ours{} (2 steps) & \cellcolor{oursrow}4.34 & \cellcolor{oursrow}3.56 & \cellcolor{oursrow}2.79 & \cellcolor{oursrow}3.02 & \cellcolor{oursrow}3.50 & \cellcolor{oursrow}2.91 & \cellcolor{oursrow}1.94 & \cellcolor{oursrow}1.83 & \cellcolor{oursrow}1.89 \\
    \midrule
    \multicolumn{11}{c}{\textit{Acceptance length $\gamma$}} \\
    \midrule
    \multirow{4}{*}{$T=0$}
      & EAGLE-3 (16) & 6.34 & 6.19 & 5.78 & \underline{5.22} & 5.51 & 5.82 & \textbf{3.52} & \textbf{3.23} & \textbf{3.84} \\
      & DFlash & 7.18 & \underline{7.13} & \underline{6.20} & 5.11 & 5.47 & 5.96 & 3.30 & 3.10 & 3.50 \\
      & \cellcolor{oursrow}\ours{} (1 step) & \cellcolor{oursrow}\underline{7.24} & \cellcolor{oursrow}6.97 & \cellcolor{oursrow}\underline{6.20} & \cellcolor{oursrow}5.18 & \cellcolor{oursrow}\underline{5.62} & \cellcolor{oursrow}\underline{6.09} & \cellcolor{oursrow}3.40 & \cellcolor{oursrow}\underline{3.17} & \cellcolor{oursrow}3.67 \\
      & \cellcolor{oursrow}\ours{} (2 steps) & \cellcolor{oursrow}\textbf{7.79} & \cellcolor{oursrow}\textbf{7.39} & \cellcolor{oursrow}\textbf{6.47} & \cellcolor{oursrow}\textbf{5.56} & \cellcolor{oursrow}\textbf{6.10} & \cellcolor{oursrow}\textbf{6.40} & \cellcolor{oursrow}\underline{3.49} & \cellcolor{oursrow}\textbf{3.23} & \cellcolor{oursrow}\underline{3.77} \\
    \addlinespace
    \multirow{4}{*}{$T=1$}
      & EAGLE-3 (16) & 6.06 & 5.37 & \textbf{4.51} & \underline{4.80} & \underline{5.18} & \textbf{5.13} & \textbf{3.21} & \textbf{3.02} & \textbf{3.26} \\
      & DFlash & \underline{6.46} & \underline{5.40} & 4.46 & 4.47 & 5.02 & 4.51 & 2.95 & 2.94 & 2.75 \\
      & \cellcolor{oursrow}\ours{} (1 step) & \cellcolor{oursrow}6.45 & \cellcolor{oursrow}5.28 & \cellcolor{oursrow}4.33 & \cellcolor{oursrow}4.55 & \cellcolor{oursrow}5.12 & \cellcolor{oursrow}4.59 & \cellcolor{oursrow}3.12 & \cellcolor{oursrow}2.82 & \cellcolor{oursrow}2.99 \\
      & \cellcolor{oursrow}\ours{} (2 steps) & \cellcolor{oursrow}\textbf{7.09} & \cellcolor{oursrow}\textbf{5.74} & \cellcolor{oursrow}\underline{4.50} & \cellcolor{oursrow}\textbf{4.89} & \cellcolor{oursrow}\textbf{5.66} & \cellcolor{oursrow}\underline{4.79} & \cellcolor{oursrow}\underline{3.13} & \cellcolor{oursrow}\underline{2.95} & \cellcolor{oursrow}\underline{3.05} \\
    \bottomrule
  \end{tabular}
  \caption{Speedup ($\times$) over autoregressive decoding with the target model (top) and acceptance length $\gamma$ (bottom) on Qwen3-4B.
  For each metric, temperature and benchmark, bold marks the best result and underline the second best.}
  \label{tab:speedup-4b}
\end{table*}

\section{Multi-Sample Drafting with Tree Verification}
\label{sec:appendix-multisample}

Because \ours{} drafts from noise, it can draft several blocks from independent noise draws in a single batched forward pass, merge them into a trie, and verify the trie as a tree in one target pass.
This is similar to what is done in DDTree~\citep{ringel_accelerating_2026}, but here every branch comes from a separate noise draw.
We ablate \ours{}'s performance at various sample sizes and tree verification sizes and report the results for Qwen3-8B at $T=0$ in 
Table~\ref{tab:multisample}.
With four samples merged into a 16-token tree---the same verification budget as a single sample---speedup rises on every benchmark, by 4--24\% (11\% on average) at one step and by 6--15\% (11\% on average) at two steps.

\begin{table*}[!h]
  \centering
  \small
  \setlength{\tabcolsep}{3.5pt}
  \begin{tabular}{@{}lccccccccccc@{}}
    \toprule
    \multirow{2}{*}{Steps} & \multirow{2}{*}{Samples} & \multirow{2}{*}{Tree size} & \multicolumn{3}{c}{Math} & \multicolumn{3}{c}{Code} & \multicolumn{3}{c}{Chat} \\
    \cmidrule(lr){4-6} \cmidrule(lr){7-9} \cmidrule(l){10-12}
    & & & GSM8K & MATH & AIME25 & MBPP & HumanEval & LCB & MT-Bench & Alpaca & Arena-Hard \\
    \midrule
    \multicolumn{12}{c}{\textit{Speedup ($\times$)}} \\
    \midrule
    \multirow{4}{*}{1}
      & 1 & 16 & 6.34 & 5.57 & 4.90 & 3.98 & 4.47 & 4.71 & 2.62 & 2.41 & 2.48 \\
      & 2 & 16 & 6.62 & 6.00 & 5.28 & 4.16 & 4.73 & 4.92 & 2.77 & 2.86 & 2.49 \\
      & 4 & 16 & 6.84 & 6.07 & 5.29 & 4.25 & 4.78 & \textbf{5.73} & 2.85 & \textbf{2.98} & \textbf{2.58} \\
      & 4 & 64 & 6.49 & 5.62 & 4.85 & 4.29 & 4.46 & 4.87 & 2.59 & 2.75 & 2.21 \\
    \addlinespace
    \multirow{4}{*}{2}
      & 1 & 16 & 6.25 & 5.52 & 4.96 & 4.01 & 4.36 & 5.03 & 2.52 & 2.59 & 2.20 \\
      & 2 & 16 & 6.51 & 5.77 & 5.12 & 4.20 & 4.63 & 5.07 & 2.74 & 2.75 & 2.49 \\
      & 4 & 16 & \textbf{6.88} & \textbf{6.31} & \textbf{5.36} & \textbf{4.39} & \textbf{4.92} & 5.33 & \textbf{2.89} & 2.97 & 2.42 \\
      & 4 & 64 & 6.61 & 5.79 & 5.06 & 4.22 & 4.71 & 5.10 & 2.72 & 2.80 & 2.16 \\
    \midrule
    \multicolumn{12}{c}{\textit{Acceptance length $\gamma$}} \\
    \midrule
    \multirow{4}{*}{1}
      & 1 & 16 & 7.97 & 6.84 & 6.01 & 4.90 & 5.50 & 6.05 & 3.22 & 2.96 & 3.04 \\
      & 2 & 16 & 8.42 & 7.44 & 6.53 & 5.16 & 5.85 & 6.27 & 3.42 & 3.54 & 3.08 \\
      & 4 & 16 & 8.66 & 7.53 & 6.55 & 5.28 & 5.92 & 7.26 & 3.52 & 3.70 & 3.15 \\
      & 4 & 64 & 9.06 & 7.84 & 6.84 & 5.89 & 6.15 & 7.07 & 3.56 & 3.78 & 3.11 \\
    \addlinespace
    \multirow{4}{*}{2}
      & 1 & 16 & 8.49 & 7.33 & 6.61 & 5.35 & 5.79 & 6.98 & 3.34 & 3.45 & 2.93 \\
      & 2 & 16 & 9.00 & 7.77 & 6.91 & 5.66 & 6.25 & 7.15 & 3.67 & 3.72 & \textbf{3.35} \\
      & 4 & 16 & 9.51 & 8.54 & 7.26 & 5.96 & 6.65 & 7.54 & 3.89 & 4.04 & 3.26 \\
      & 4 & 64 & \textbf{9.95} & \textbf{8.68} & \textbf{7.69} & \textbf{6.25} & \textbf{7.00} & \textbf{7.99} & \textbf{4.04} & \textbf{4.15} & 3.28 \\
    \bottomrule
  \end{tabular}
  \caption{Multi-sample drafting with \ours{} on Qwen3-8B at $T=0$: speedup ($\times$) over the target and acceptance length $\gamma$ for various numbers of steps, sample counts, and tree verification sizes.
  For each metric and benchmark, bold marks the best result.
  Calculated on one H100 SXM with 32 prompts per benchmark (30 for AIME25).}
  \label{tab:multisample}
\end{table*}

\section{Additional conditional acceptance results}
\label{sec:appendix-cond}

\begin{figure*}[t]
  \centering
  \includegraphics[width=\textwidth]{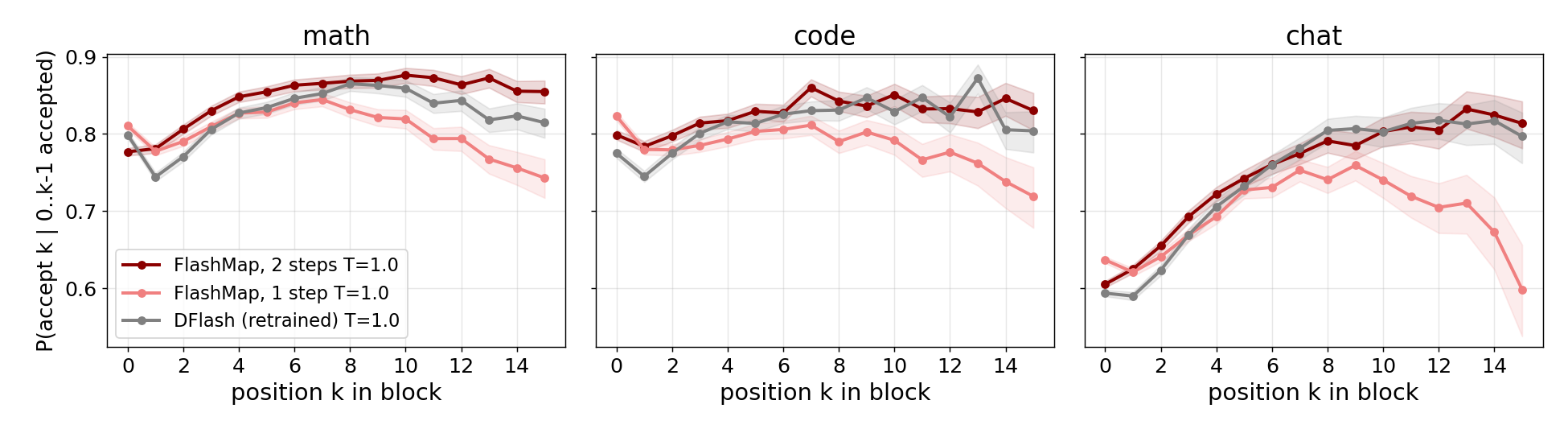}
  \caption{Same as Figure \ref{fig:decay8b} but at $T=1$.}
  \label{fig:decay8b-t1}
\end{figure*}

\begin{figure*}[t]
  \centering
  \includegraphics[width=\textwidth]{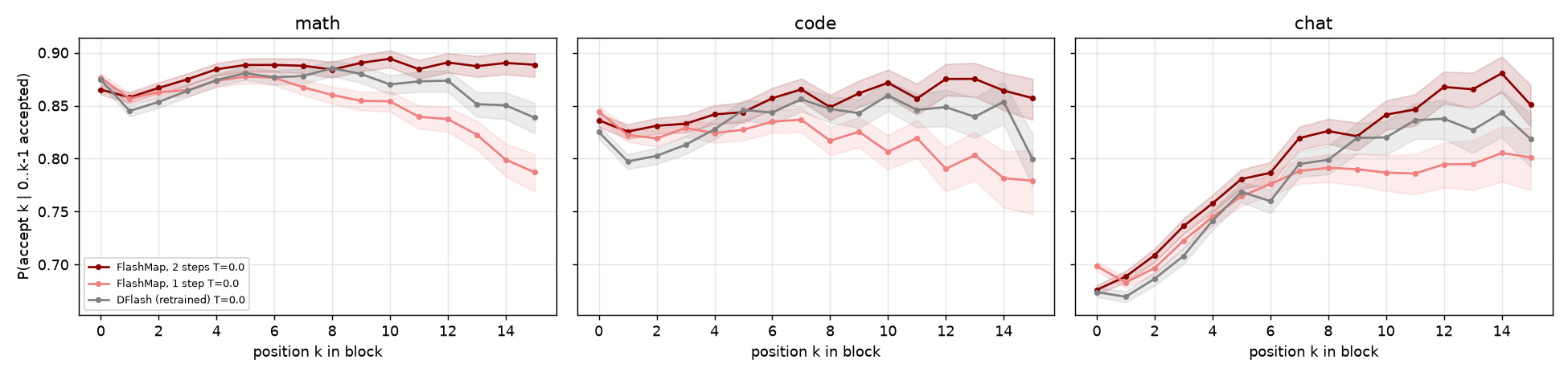}
  \includegraphics[width=\textwidth]{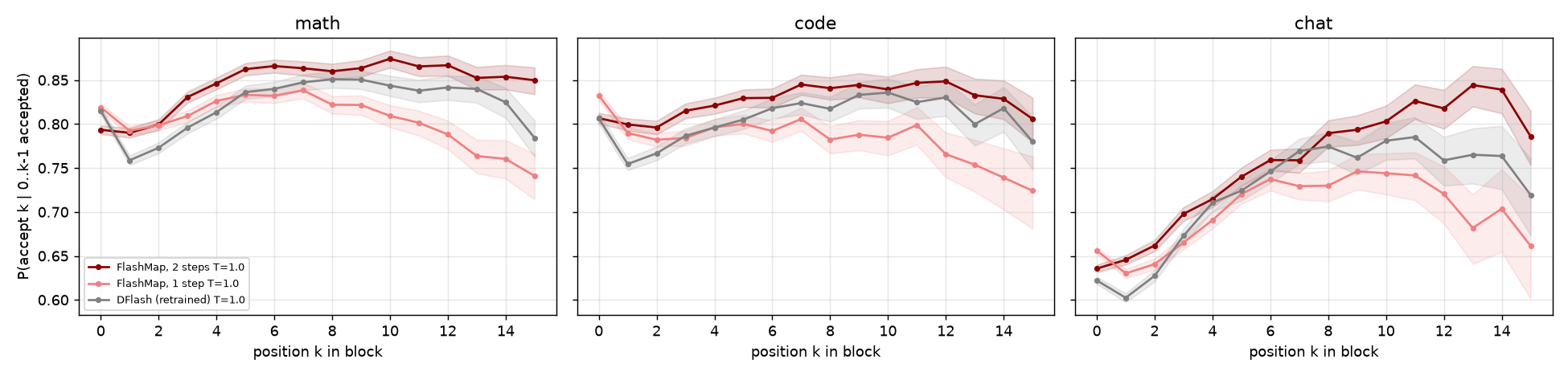}
  \caption{Same as Figure \ref{fig:decay8b}, but for Qwen3-4B at $T=0$ (top row) and $T=1$ (bottom row).}
  \label{fig:decay4b}
\end{figure*}

We show conditional acceptance curves for Qwen3-8B at $T=1$ in Figure \ref{fig:decay8b-t1}.
This is the counterpart to the $T=0$ results in Figure \ref{fig:decay8b}.
Similarly, we show conditional acceptance curves for Qwen3-4B at both $T=0$ and $T=1$ in Figure \ref{fig:decay4b}.

\section{Shuffle test on Qwen3-4B}

Table \ref{tab:shuffle4b} shows the results of the shuffle test, as described in Section \ref{sec:shuffle}, with Qwen3-4B as the target model.
These results are analogous to the ones in Table \ref{tab:shuffle8b}, and the conclusions are largely the same. 

\begin{table*}[!ht]
  \centering
  \small
  \setlength{\tabcolsep}{5pt}
  \begin{tabular}{@{}lccccccccc@{}}
    \toprule
    & \multicolumn{3}{c}{Math} & \multicolumn{3}{c}{Code} & \multicolumn{3}{c}{Chat} \\
    \cmidrule(lr){2-4} \cmidrule(lr){5-7} \cmidrule(l){8-10}
    Drafter & Shared & Shuffled & $\Delta$ & Shared & Shuffled & $\Delta$ & Shared & Shuffled & $\Delta$ \\
    \midrule
    DFlash        & 7.44 & 7.44 & 0.00 & 5.74 & 5.74 & 0.00 & 2.88 & 2.88 & 0.00 \\
    \addlinespace
    Flow (1 step) & 6.74 & 6.74 & 0.00 & 5.20 & 5.20 & 0.00 & 2.78 & 2.78 & 0.00 \\
    Flow (2 steps)          & 7.70 & 7.25 & 0.46 & 6.12 & 5.79 & 0.34 & 3.04 & 2.88 & 0.16 \\
    Flow (16 steps)         & 8.05 & 7.22 & 0.83 & 6.47 & 5.84 & 0.62 & 3.17 & 2.88 & 0.29 \\
    \addlinespace
    \ours{} (1 step) & 7.07 & 6.71 & 0.36 & 5.60 & 5.45 & 0.15 & 2.86 & 2.77 & 0.08 \\
    \ours{} (2 steps) & 7.74 & 6.87 & 0.87 & 6.25 & 5.72 & 0.53 & 3.08 & 2.82 & 0.26 \\
    \bottomrule
  \end{tabular}
  \caption{Same as Table \ref{tab:shuffle8b} but for Qwen3-4B.}
  \label{tab:shuffle4b}
\end{table*}

\section{Implementation and compute details}
\label{sec:appendix-impl}

\paragraph{Model sizes.}
On Qwen3-8B, \ours{} has 500M trainable parameters, DFlash has 1049M and EAGLE-3 has 302M.
On Qwen3-4B, \ours{} has 475M trainable parameters, DFlash has 537M, and EAGLE-3 has 149M.

\paragraph{Compute.}
We train on 4 NVIDIA H200 SXM5 nodes with 8 GPUs each (total of 32 GPUs). 
Training one \ours{} flow teacher took ${\sim}25$ hours for Qwen3-8B and ${\sim}17$ hours for Qwen3-4B.
Distillation into a flow map took ${\sim}9$ hours.
The retrained DFlash took ${\sim}27$ hours for Qwen3-8B and ${\sim}18$ hours for Qwen3-4B.
The retrained EAGLE-3 took ${\sim}21$ hours for Qwen3-8B and ${\sim}16$ hours for Qwen3-4B.
Across all target models, drafters, and ablations, the total is about 4600 GPU hours.
Speedups are measured on one NVIDIA H200 NVL GPU (one H100 SXM for Table~\ref{tab:multisample}).

\paragraph{Software.}
We implement all drafters in DeepSpec 005e03b with PyTorch and evaluate with SGLang 0.5.20 at concurrency 1, with 2048 max new tokens and thinking mode off.

\paragraph{Evaluation data.}
Licenses: GSM8K, MATH, LiveCodeBench and HumanEval (MIT); AIME25, MBPP, MT-Bench, AlpacaEval, and Arena-Hard (Apache 2.0).

\paragraph{Variance.}
All speedups and acceptance lengths are from a single run per configuration. At $T=0$, re-running a configuration on a different GPU changed acceptance length by at most about 1\%. Conditional acceptance curves show 95\% Wilson intervals.
\end{document}